\documentclass[letterpaper]{article} 
\usepackage[usenames, dvipsnames, table]{xcolor} 
\usepackage{aaai2026}  
\usepackage{times}  
\usepackage{helvet}  
\usepackage{courier}  
\usepackage[hyphens]{url}  
\usepackage{graphicx} 
\usepackage{natbib}  
\usepackage{caption} 
\usepackage{algorithm}
\usepackage[noend]{algpseudocode}

\usepackage{newfloat}
\usepackage{listings}
\DeclareCaptionStyle{ruled}{labelfont=normalfont,labelsep=colon,strut=off} 
\floatstyle{ruled}
\newfloat{listing}{tb}{lst}{}
\floatname{listing}{Listing}
\usepackage{color}
\definecolor{tabfirst}{rgb}{1, 0.7, 0.7} 
\definecolor{tabsecond}{rgb}{1, 0.85, 0.7} 
\definecolor{tabthird}{rgb}{1, 1, 0.7} 
\definecolor{Gray}{gray}{0.85}

\RequirePackage{xspace}
\makeatletter
\DeclareRobustCommand\onedot{\futurelet\@let@token\@onedot}
\def\@onedot{\ifx\@let@token.\else.\null\fi\xspace}

\def\eg{\emph{e.g}\onedot}

\makeatother

\usepackage{amsmath,amsfonts,bm}

\def\eqref#1{equation~\ref{#1}}

\def\1{\bm{1}}

\def\rvc{{\mathbf{c}}}
\def\rvd{{\mathbf{d}}}

\def\rvr{{\mathbf{r}}}

\def\rvx{{\mathbf{x}}}

\def\mB{{\bm{B}}}

\def\mD{{\bm{D}}}

\def\mI{{\bm{I}}}

\def\mP{{\bm{P}}}

\DeclareMathAlphabet{\mathsfit}{\encodingdefault}{\sfdefault}{m}{sl}
\SetMathAlphabet{\mathsfit}{bold}{\encodingdefault}{\sfdefault}{bx}{n}

\def\sR{{\mathbb{R}}}

\newcommand{\E}{\mathbb{E}}

\usepackage[capitalize]{cleveref}
\usepackage{adjustbox}
\usepackage{booktabs}
\usepackage{multirow}
\usepackage[caption=false,font=normalsize]{subfig}

\title{Surface-Based Visibility-Guided Uncertainty for\\Continuous Active 3D Neural Reconstruction}
\author {
    Hyunseo Kim\textsuperscript{\rm 1},
    Hyeonseo Yang\textsuperscript{\rm 1},
    Taekyung Kim\textsuperscript{\rm 2}, 
    Yoonsung Kim\textsuperscript{\rm 1},
    Minsu Lee\textsuperscript{\rm 3},
    Jin-Hwa Kim\textsuperscript{\rm 1,\rm 2}\equalcontrib,
    Byoung-Tak Zhang\textsuperscript{\rm 1}\equalcontrib,
}
\affiliations {
    \textsuperscript{\rm 1}Artificial Intelligence Institute, Seoul National University\\
    \textsuperscript{\rm 2}NAVER AI Lab\\
    \textsuperscript{\rm 3}Sungshin Women's University\\
     \{hskim, hsyang, yskim, btzhang\}@bi.snu.ac.kr,
     \{taekyung.k, j1nhwa.kim\}@navercorp.com, 
     mslee@sungshin.ac.kr
}

\begin{document}

\maketitle

\begin{abstract}
View selection is critical in active 3D neural reconstruction as it impacts the contents of training set and resulting final output quality.
Recent view selection strategies emphasize the visibility when evaluating model uncertainty in active 3D reconstruction.
However, existing approaches estimate visibility only after the model fully converges, which has confined their application primarily to non-continuous active learning settings. 
This paper proposes Surface-Based Visibility field (SBV) that successfully estimates the visibility-guided uncertainty in continuous active 3D neural reconstruction. 
During learning neural implicit surfaces, our model learns rendering uncertainties and infers surface confidence values derived from signed distance functions. 
It then updates surface confidences using a voxel grid, robustly deducing the surface-based visibility for uncertainties.
This approach captures uncertainties across all regions, whether well-defined surfaces or ambiguous areas, ensuring accurate visibility measurement in continuous active learning. 
Experiments on benchmark datasets—Tanks and Temples, BlendedMVS, Blender, DTU—and the newly proposed imbalanced viewpoint dataset (ImBView) show that view selection based on SBV-guided uncertainty improves performance by up to 11.6\% over existing methods, highlighting its effectiveness in challenging reconstruction scenarios.
\end{abstract}

\begin{links}
    \link{Code}{https://github.com/hskAlena/Surface-Based-Visibility}
\end{links}

\section{Introduction}
\label{sec:intro}
Active 3D reconstruction has been extensively studied to reduce data acquisition and computational costs in multi-view 3D reconstruction \cite{chen2011active, isler2016information}.
For view acquisition, 3D reconstruction models assess uncertainties within a scene and evaluate the extent to which these uncertainties can be reduced by acquiring additional views.
The expected reduction in uncertainty is referred to as the expected information gain (IG), and the candidate view with the maximum expected IG is termed the next-best view (NBV). 
For IG computation, visibility in reconstructed scenes is crucial and has been extensively addressed in prior work.

\begin{figure}[!t]
     \centering
     \includegraphics[width=1.\linewidth]{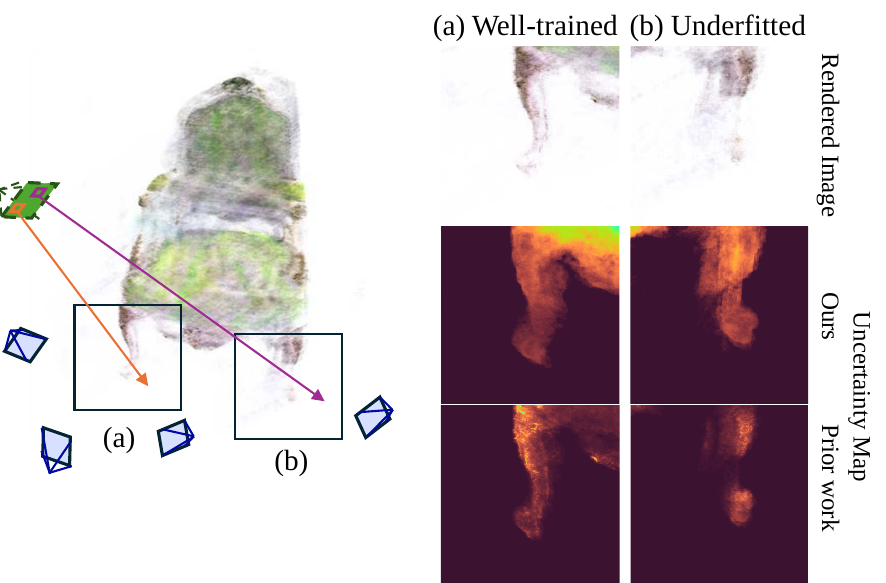}
     \caption{
     \textbf{Problem statement.}
In continuous active 3D neural reconstruction, a model is initialized with a small number of views (blue) and evaluates the information gain of a candidate view (green) for next-best view selection.
Regions trained with low data (b) exhibit lower convergence than those trained with high data (a).  
Prior methods often fail to capture uncertainties in underfitted regions with low volume densities, while our method performs well in both regions. 
     }
     \label{fig:framework}
 \end{figure} 
 
Prior active 3D neural reconstruction methods typically employ volume density or voxel occupancy to account for visibility. 
They \cite{xue2024nvf} compute the visibility using volume density after a substage of the training has converged. 
However, these post hoc visibility estimation frameworks are ill-suited for continuous active 3D neural reconstruction, where visibility must be evaluated in real time during ongoing model updates.

To address this challenge, we propose surface-based visibility field (SBV) that enables visibility inference for uncertainty in continuous learning. 
Our model simultaneously learns a neural implicit surface while estimating rendering uncertainty of a scene, and updates confidence values for estimated surfaces using voxel grids during training. 
The adoption of surface confidence values, robustly updated using a voxel grid, facilitates the stable computation of surface-based visibility in continuous learning scenarios. 
Leveraging surface-based visibility-guided uncertainty allows the model to handle complex object occlusions during continuous learning effectively, which was validated on four benchmark datasets.
To visually analyze different NBV selection strategies, we constructed a toy dataset with imbalanced viewpoints, explicitly restricting the accessible information from each view.

\textbf{Contributions.} In summary, the contributions of this study are as follows:
\begin{itemize}
\item 
We propose surface-based visibility field, SBV, which enables the estimation of visibility for uncertainties in continuous active 3D neural reconstruction.
\item 
To the best of our knowledge, this is the first study to address the potential inaccuracy of volume rendering in assessing the visibility of uncertain regions during continuous active learning.
\item 
The proposed SBV-based view selection method outperforms other approaches on four standard benchmarking datasets, exhibiting consistent performance improvements of up to 11.6\% in image rendering across diverse scene scales and complexities.
\item  
We introduce ImBView, a new toy dataset featuring imbalanced viewpoint distributions that enables clear analysis on view selection strategies.  
\end{itemize}

\section{Related Work}
\label{related_work}

\subsection{Active 3D Neural Reconstruction}
\label{sec:related_uncertainty}
Active 3D neural reconstruction reduces the significant costs associated with conventional 3D neural reconstruction and allows reconstruction in unknown environments where training data may be scarce or unavailable. 
Early studies can be broadly categorized into those utilizing supplementary depth data for large-scale scene reconstruction \cite{ran2023neurar,yan2023active} and those relying solely on RGB data to reconstruct specific objects \cite{goli2023bayes,pan2022activenerf}. 
Both categories implement active view planning or selection by quantifying uncertainty or information within their respective models. 
This study specifically addresses the task of reconstructing individual objects by using only RGB inputs, without depth data.

The methods differ mainly in their definitions of model uncertainty, which determines the resulting view selection and final reconstruction quality. 
Some approaches define uncertainty with respect to the volume density distribution along rays \cite{UncertaintyGuidedPolicy}, while others estimate it based on the variance of color distributions at individual points \cite{pan2022activenerf}. 
In methods that employ explicit data structures for rendering, view selection strategies often rely on entropy \cite{zhan2022activermap} or Fisher information \cite{Jiang2023FisherRF} within the data representation. 
Despite these advances, none explicitly address the crucial aspect of visibility in measuring uncertainty.

Since the neural network implicitly encodes 3D scene information, visibility within the reconstructed scene is crucial for filtering uncertainty that directly impacts model output quality.
Thus, accurate modeling and accounting for visibility are critical for informed view selection and robust reconstruction performance.

\subsection{Visibility-Guided Uncertainty for Active 3D Reconstruction}
\label{sec:related_explicit}
In active 3D reconstruction, considering visibility allows for distinguishing uncertainty on the object surface, which directly influences the model output. 
This technique facilitates data-efficient reconstruction and reduces convergence time. 
However, in active 3D reconstruction using volumetric density fields, reliably estimating the visibility during training remains challenging. 
Consequently, methods like NVF \cite{xue2024nvf} estimate visibility-guided uncertainty only after completing model training on all accessible inputs. 
Additional views are then selected based on this visibility-guided uncertainty, and model training is resumed.

Conversely, in the domain of active 3D reconstruction employing voxel grids \cite{isler2016information}, additional views can be continuously selected during reconstruction. 
These methods evaluate the informativeness of views by weighting the uncertainty on regions near observed surfaces. 
Nevertheless, the determination process and reliability of surfaces differ between voxel-based and neural reconstruction in continuous active learning, which complicates the application of successful voxel-based approaches to neural methods.

To address this, we propose the concept of surface confidence for inferring surface-based visibility in continuous active 3D neural reconstruction. 
We further introduce a voxel grid-based update strategy for fast uncertainty computation, inspired by Instant NGP \cite{muller2022instant} and Nerfacc \cite{li2023nerfacc}.

\begin{figure}[!t]
\centering
     \includegraphics[width=1.\linewidth]{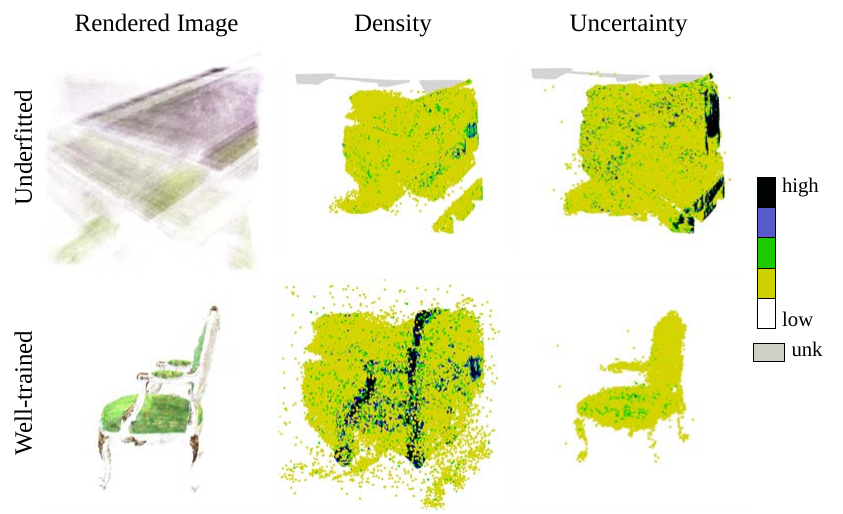}
    \caption{
    Analysis of density-based visibility in continuous active 3D neural reconstruction.
Volume density, visualized based on bitfield values,
grows from low (yellow) to high (black) as additional views are included and training progresses, while uncertainty decreases from high to low.
Gray regions represent unobservable areas from limited training views in the early stages of training. 
     }
     \label{fig:Ause}
 \end{figure}

\section{Estimation of Visibility in Continuous Active 3D Neural Reconstruction}
\label{motivation}
We analyzed the intermediate results of a volumetric density-based model to assess the accuracy of visibility estimation in continuous active 3D reconstruction.
In \cref{fig:Ause}, we present volume-rendered images along with Trimesh~\cite{trimesh}-visualized volume density and uncertainty bitfields for well-trained and underfitted stage of models.
The distribution of the volume density is visualized using Trimesh to provide a clear 3D spatial representation, avoiding the limitations of 2D volume rendering.
Notably, rendering uncertainty-defined as the variance of the color distribution-is invariant with respect to ray direction and is therefore visualized using bitfields.

In the early stages of training, uncertainty decreases rapidly in areas inferred as empty spaces while remaining high near target objects. 
To enhance the reconstruction of target objects, the model must prioritize selecting views that include visible regions with high uncertainty.
However, IG quantification using volume rendering fails to account for high uncertainty in regions with underestimated volume densities.
Consequently, informative candidate views during early training are often excluded from view selection or are only included after extensive training, thus reducing their potential contribution to the overall reconstruction performance.

After sufficient training, low uncertainty and high volume densities are observed near the target object. However, as shown in \cref{fig:Ause}, numerous floaters with low volume densities are present around the target. These artifacts, commonly reported in prior NeRF reconstructions~\cite{warburg2023nerfbusters, philip2023floaters}, arise from sparse input images.
While neural implicit surfaces \cite{wang2021neus} can mitigate floaters using surface constraints, their complete elimination is neither achievable nor the focus of this study. 
IG computation using volume rendering also fails to recognize the uncertainty of floaters, which could pose challenges if the uncertainty of floaters increases with the introduction of additional views.

This analysis does not argue against the use of volume rendering for training models to predict scene uncertainties. 
Instead, it highlights that employing volume rendering to determine the \textbf{visibility} of uncertain regions during the \textbf{early, underfitted} training stages may lead to inaccurate uncertainty estimates.
The introduction of SBV addresses this limitation by enabling more accurate IG computation during underfitted training stages. SBV achieves this by quantifying uncertainty based on surface confidence, thereby accounting for both complete and incomplete inferred surfaces.

\section{Method}
\label{methods}
In this section, we propose a new information gain (IG) formulation for selecting the next-best view (NBV). 
Our IG formulation incorporates both uncertainty and surface confidence to accurately infer the surface-based visibility (SBV) of uncertain regions.

First, the signed distance function (SDF) value and rendering uncertainty of 3D points are estimated by extending NeuS \cite{wang2021neus} with an uncertainty prediction branch \cite{pan2022activenerf}. 
We then introduce a voxel grid for robust update of surface confidence and efficient IG computation.
This approach enables accurate surface-based visibility inference for uncertain regions and effective view selection, even when the model is underfitted.
Lastly, we introduce a method for selecting multiple NBVs using SBV-guided uncertainty.

\subsection{Preliminary: Neural Implicit Surfaces}
\label{sec-prelim_neus}
NeuS \cite{wang2021neus} utilizes a neural signed distance function (SDF) to encode scene geometry. 
For a 3D point's position $\rvx$ and viewing direction $\rvd$, the neural implicit surface network (NeuS) represents the scene geometry through the SDF $g(\rvx)$ and color $\rvc$ ($F_\Theta : (\rvx, \rvd) \rightarrow (g(\rvx), \rvc)$).
The zero-level set of the SDF defines the surface $\mathcal{S}$ of an object:
\begin{equation}
 \mathcal{S} = \{ \rvx \in \sR^3 | g(\rvx)=0 \}.
 \label{eq:sdf_def}
\end{equation}
 NeuS introduces the S-density field $\phi_s(g(\rvx))$
to integrate the SDF with volume rendering \cite{levoy1990volumerender}.
Here, $\phi_s(x)$ is the derivative of the Sigmoid function 
$\Phi_s(x)= (1+ e^{-sx} )^{-1}$, which forms a zero-centered unimodal density distribution.
The standard deviation of $\phi_s$ is $1/s$, which approaches zero as training converges, with $s$ being a learnable parameter. 
The value $1/s$ can also be interpreted as the point sampling step size, indicating that the network predicts a more precise and thinner surface as training progresses. Thus, when the network is underfitted, $1/s$ remains large. 
To address general volume rendering scenarios, NeuS introduces opaque density $\rho$, analogous to NeRF's \cite{mildenhall2021nerf} volume density $\sigma$, and a discrete opacity value $\alpha$:
\begin{align}
\alpha_j  = 1- \exp \left( -\int_{t_j}^{t_{j+1}} \rho(t) dt \right), 
\label{eq:neus_alpha}  
\\
\rho(t_j) = \max \left( \frac{-\frac{d\Phi_s}{dt} (g(\rvr_i(t_j)))}{\Phi_s(g(\rvr_i(t_j)))}, 0 \right)
 \label{eq:neus_rho}
\end{align}
where $N$ points are sampled along the $i$-th ray $\rvr_i(t_j)$.
After calculating the weight $\omega_i = \alpha_i\prod_{j=1}^{i-1} (1-\alpha_j)$, the volume rendering equation is given as follows:
\begin{align}
\hat{C}(\rvr) = \int_{t_n}^{t_f} \omega(t) \rvc(\rvr(t), \rvd) dt, 
\label{eq:nerf_cont_color}
\end{align}
where the integration of color values occurs along the ray from the near plane $t_n$ to the far plane $t_f$.
The final loss function comprises both color loss and Eikonal \cite{gropp2020implicit} regularization losses.
 The color loss, as shown in \cref{eq:neus_loss}, is averaged over $M$ camera rays, where $C(\rvr)$ and $\hat{C}(\rvr)$ represent the ground truth and predicted colors for ray $\rvr$, respectively. 
The Eikonal loss ensures that the gradients of the SDF, $\nabla g(\rvr_i (t_j))$, have a unit 2-norm, regulated by the parameter $\lambda$. The average Eikonal loss is computed over the number of sampled points $N$ as follows: 
\begin{align}
\mathcal{L_\textit{s}} &=  
\frac{1}{M} \sum_{i=1}^{M} \Big[ \|(\hat{C}(\rvr_i)- C(\rvr_i) \|_1 
\label{eq:neus_loss}  \\\nonumber
 &+ \frac{\lambda}{N} \sum_{j=1}^{N} (\| \nabla g(\rvr_i (t_j))\|_2 - 1)^{2} \Big]
\end{align}

\subsection{Estimation of Uncertainty}
\label{method:uncertain_neus}
We first define the uncertainty of a scene for a neural implicit surface network. 
To estimate the rendering uncertainty, we model the color of each 3D point as a Gaussian distribution (\cref{gauss}), with the color variance representing the uncertainty. 
\begin{equation}
c(\rvr(t)) \sim \mathcal{N}(\bar{c}(\rvr(t)), \beta^2 (\rvr(t))) 
 \label{gauss}
\end{equation}
The neural implicit surface network is extended to incorporate color as a Gaussian distribution, defined as: 
$F_\Theta : (\rvx, \rvd) \rightarrow (g(\rvx), \bar{c}, \beta^2)$, where $\bar{c}$ and $\beta^2$ represent the mean and variance of the color $c$, respectively.
Notably, the color variance $\beta^2$, which reflects the inconsistency in color when the viewpoint changes, is independent of the view direction $\rvd$.
Finally, we define the uncertainty loss, $\mathcal{L}_{\textit{u}}$, as the negative log-likelihood of the predicted color distribution. This loss is used to optimize the mean and variance of each ray as follows:
\begin{equation}
\mathcal{L}_{\textit{u}} = \frac{1}{M} \sum^{M}_{i=1} \left( \frac{\| \bar{\mathcal{C}}(\rvr_i) -  C(\rvr_i) \| ^{2}_{2}}{2\mathcal{B}^2(\rvr_i)} + \frac{\log \mathcal{B}^2 (\rvr_i)}{2} \right)
\label{eq:uncert_color_loss}
\end{equation}
where $C(\rvr_i)$ represents the ground truth color from the training image, and $M$ is the number of camera rays. The mean $\bar{\mathcal{C}}(\rvr)$ and variance $\mathcal{B}^2(\rvr)$ of a ray are computed using volume rendering \cite{levoy1990volumerender}.
To incorporate the rendering uncertainty along with surface information (\cref{eq:neus_loss}), the final loss function is defined as $\mathcal{L} =  \mathcal{L}_\textit{s} + \omega \mathcal{L}_\textit{u}$, where $\omega$ is a hyperparameter that balances the two components.

\begin{figure}[!t]
\centering
\includegraphics[width=0.85\linewidth]{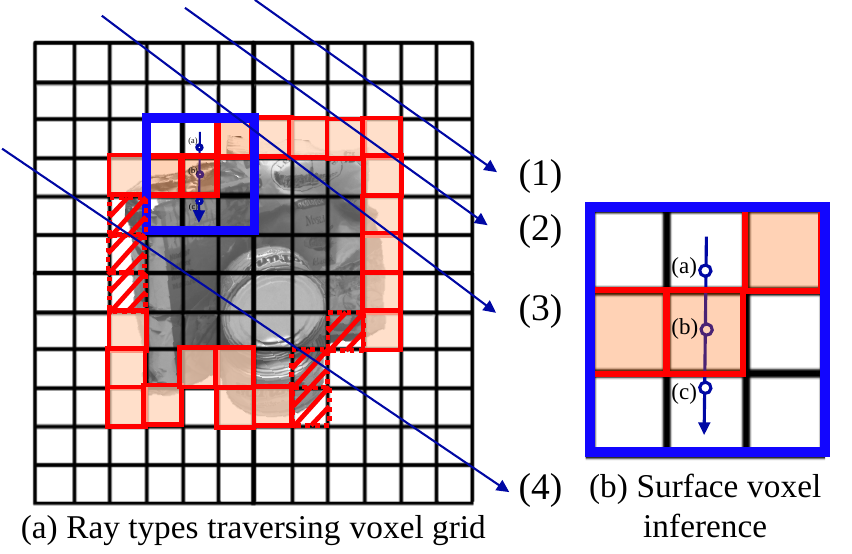}
\caption{
Scenarios for uncertainty estimation using surface-based visibility.
Voxels containing inferred surfaces are highlighted in red, while voxels marked with diagonal red lines indicate low surface confidence in the current training stage but potentially form part of surface. 
  \textbf{(a)}: Rays traversing no visible surfaces (rays (1) and (4)) versus visible surfaces (rays (2) and (3)) exhibit distinct surface-based visibilities.
  \textbf{(b)}: Demonstration of inferring surface confidence using estimated SDF. 3D points (a), (b), and (c) are points on same camera ray. Distance between (a) and (c) is sampling step size $1/s$ inferred from neural implicit surface network, with (b) being the midpoint.}
\label{fig:uncertain_grid}
\end{figure}

\subsection{Preliminary: Information Gain in Voxel Grids}
In voxel-based NBV selection methods, the information within a voxel grid is defined as the entropy $I(x)$ of voxel occupancy probability~\cite{isler2016information}. 
When the voxel grid is initialized, the occupancy probability $p(x)$ of a voxel $x$ is set to 0.5, as the occupancy status of the voxel is yet to be determined. 
Therefore, the initialized voxel has the highest entropy (expected information).
The entropy $I(x)$ is formulated as follows:
\begin{equation}
    I(x) = - p(x) \log p(x) - (1-p(x)) \log(1-p(x)).
\label{dens_ent}
\end{equation}
The IG in a voxel grid is defined as the sum of the expected information enclosed in the voxels visible from a specific view \cite{thrun2002probabilistic}. 
Let $\mathcal{R}_v$ represent the set of camera rays traversing the voxel grid for camera view $v$, and $\mathcal{X}_r$ denote the set of voxels through which the ray $r$ passes. 
The IG $G(v)$ for a camera view $v$ is expressed as follows:
\begin{equation}
    G(v) = \frac{1}{n} \sum_{ r \in \mathcal{R}_v} \sum_{ x\in \mathcal{X}_r} I(x).
\label{dens_ent_infogain}
\end{equation}
where $n$ is the total number of traversed voxels.

\subsection{Information Gain under Surface-Based Visibility}
\label{method:infogain}
SBV is updated using a voxel grid that encapsulates both uncertainty and surface confidence. 
So, the IG quantification of SBV-guided uncertainty is formulated based on the voxel-based IG quantification formulation $G(v)$ in \cref{dens_ent_infogain}.
To apply this formulation, we convert uncertainty into entropy, as the entropy of the voxel grid serves as the foundation for computing $G(v)$.
Since the color is represented using a Gaussian probability distribution, the entropy $H$ of the color is defined in terms of the color variance $\beta^2$, as described in \cite{cover1999elements}, as follows: 
\begin{align}
H(c(\rvr(t))) &= - \E [\log \mathcal{N} (\bar{c}(\rvr(t)), \beta^2 (\rvr(t)))]  \label{color_ent1} \\ 
&= \frac{1}{2} \log(2 \pi \beta^2) + \frac{1}{2}
\label{color_ent}
\end{align}
 Empirically, $\beta^2 (\rvr(t))$ takes values in the range of 0 to 1. Therefore, we initialize the rendering uncertainty value of a voxel grid to 1, ensuring the entropy is maximized at the start. 
 To stabilize the assigned uncertainty values as the network weights evolve during training~\cite{li2023neuralangelo}, we implement a strategic update mechanism.
When updating the uncertainty, the voxel grid selects the minimum value between the previous uncertainty (increased by a factor of 1.05) and the newly predicted uncertainty.

After updating scene uncertainty using the voxel grid, surface confidence scores are updated to vary the IG formulation based on surface-based visibility. 
To perform voxel-level surface detection, as illustrated in \cref{fig:uncertain_grid}b, we sample three points along a camera ray: the voxel center (b) with added random noise, and two adjacent points (a, c). 
A surface is detected if the SDF values of points (a) and (c) have opposite signs, resulting in a negative SDF product.
Surface confidence score is estimated as a binary value: 1 if the SDF product is negative, and 0 otherwise. Then, it is stored in each voxel and updated as $\max(\texttt{previous score} \times \texttt{decay rate}, \texttt{current estimation})$, using a decay rate of 0.95.
The decay rate ensures robust surface estimation, compensating for the random noise in the voxel center (b) sampling.
A voxel is classified as a surface voxel if its confidence score exceeds the threshold of 0.8. 

Finally, as depicted in \cref{fig:uncertain_grid}a, we formulate the IG quantification of SBV-guided uncertainty. 
A region is designated as ambiguous when camera rays do not intersect surface voxels while traversing voxel grids (\eg, camera rays (4)). 
If camera rays intersect surface voxels, the region is considered to have well-defined surfaces (\eg, camera rays (2) and (3)). 
The camera ray (1), which traverses a vacant space, does not encounter any surface voxel, similar to camera ray (4). The uncertainty of vacant spaces is estimated to be low, even during the early training stages.
Therefore, we treat camera rays (1) and (4) as equivalent cases with low surface confidence and compute the IG by summing the color entropy of the traversed voxels.
For camera rays that intersect well-defined surfaces, we compute the IG using only the color entropy of the surface voxel, $H(c(x))$, as derived from 
\cref{color_ent}. In summary, our IG formulation, $G_s(v)$, is:
\begin{align}
    G_s(v) = \frac{1}{N} \sum_{ r \in \mathcal{R}_v} \sum_{ x\in \tilde{\mathcal{X}}_r} H(c(x))
\label{color_ent_infogain}
\end{align}
Here, $\tilde{\mathcal{X}}_r$ represents $\mathcal{X}_r \cap \mathcal{S}$ when the camera ray $r$ intersects surface voxels; otherwise, it denotes $\mathcal{X}_r$, the set of voxels traversed by the ray. $N$ is defined as $\sum_{r \in \mathcal{R}_v} |\tilde{\mathcal{X}}_r|$.

\subsection{Multiple Next-Best View Selection}
\label{sec:method_multipleNBV}
Selecting multiple NBVs is essential when acquiring diverse perspectives while being limited by computational resources and time. 
Simply selecting the top $k$ candidates based solely on their IG values often leads to the selection of similar views from a limited region of the camera sphere, causing redundancy and limiting the diversity of information.

To address this issue, we introduce a selection strategy ensuring that the selected $k$ views are sufficiently spaced apart. The procedure for selecting these candidates is outlined in \cref{alg:cand_k_view} in the supplementary.
The main goal is to select views with high IG values while maintaining a minimum distance $\tau$ from all previously incorporated train views.
A view is considered eligible if its distance from all train views exceeds $\tau$ and it has the highest IG value among the qualified candidates. 
Empirically, $\tau$ is initialized to 1.732. If no candidate satisfies the distance threshold, $\tau$ is decreased by multiplying it by a decay factor of 0.95.

\section{Experiment}
\label{Experiments}
\textbf{Implementation details.} We evaluated continuous active 3D neural reconstruction performance in two settings: two-image and ten-image settings.
For DTU, Blender, and ImBView, we used the two-image setting, while for TanksAndTemples and BlendedMVS, we used the ten-image setting due to their complexity and large-scale nature.
In both settings, training begins with a training set comprising two or ten pre-designated images. 
After a predefined number of iterations, the model selects two or ten NBV images from the candidate set to add to the training set, subsequently removing them from the candidate set. 
This process is repeated four times. Then, view selection is stopped, and training continues until the specified number of iterations is reached.
Therefore, 10 or 50 images are used for training in the two- or ten-image settings, respectively.
Additionally, we evaluated Blender and DTU on other settings (one- or four-image settings).
The results and detailed experimental settings can be found in the supplementary material.
 
We used the Adam optimizer \cite{kingma2014adam} with a learning rate of 0.0005. The network was implemented in PyTorch and trained on a single NVIDIA RTX 2080 GPU for approximately 10 hours until convergence. Detailed hyperparameter settings are also included in the supplementary material.

\textbf{Comparing methods.} We compared our method against five types of NBV selection methods in continuous active 3D neural reconstruction: Random, furthest view sampling (FVS) \cite{Xiao:CVPR24:NeRFDirector}, Entropy from ActiveRMAP \cite{zhan2022activermap}, ActiveNeRF \cite{pan2022activenerf}, and FisherRF \cite{Jiang2023FisherRF}. 
Note that all compared methods were re-implemented on the NeuS framework to ensure a fair comparison. To distinguish our implementations of ActiveNeRF and FisherRF from the original papers, we refer to them as ActiveNeRF* and FisherRF*, respectively. 
We renamed the NBV selection method from ActiveRMAP as 'Entropy' because we exclusively implemented its IG computation using entropy, while omitting its original mapping and planning strategies.
Further details are provided in the supplementary material.

\subsection{Information Gain Visualization}
\begin{figure}[!t]
     \centering
\includegraphics[width=1.\linewidth]{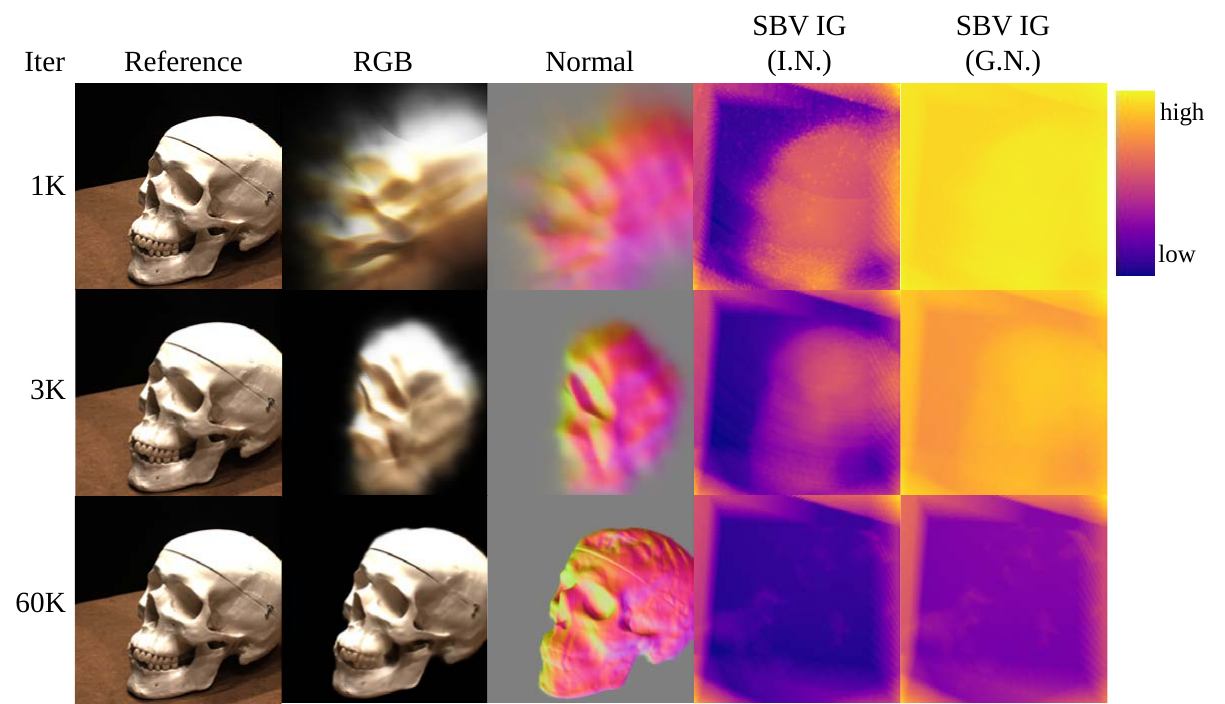}
     \caption{
     Visualization of surface-based visibility-guided uncertainty as training progresses.
     IG quantified with SBV is normalized using two methods: intra-image (I.N.) and global normalization (G.N.).
Former normalizes image using its minimum and maximum IG values. In contrast, latter normalizes images using minimum of IG from 60K iterations and maximum of IG from 1K iterations. 
}
\label{fig:IG_iter}
 \end{figure}

We visualized the reduction in IG computed using SBV across test views during model training. 
The experiments were conducted on the DTU dataset in a two-image setting, where training begins with two initial views and incrementally adds two views at iterations 1K, 2K, 3K, and 4K through view selection.
\cref{fig:IG_iter} illustrates the results at various stages of training. At 1K and 3K iterations, the novel view synthesis outcomes and IG visualizations were generated using the model trained on the two and six views, respectively, just before the subsequent view selection.
The 60K iteration results reflect the model's final performance after training on all ten views acquired during the view selection process.

The normal map is derived from the gradient of the signed distance function (SDF). In regions without surfaces, the gradient typically points toward the nearest surface. However, in the datasets used in this study, where backgrounds are either white or masked to black during training, the SDF gradients in background regions converge to $(0,0,0)$. When visualized using the $(\texttt{normal}+1)/2$ transformation, these empty spaces appear as gray, corresponding to RGB values of $(0.5, 0.5, 0.5)$.
Colored regions in the normal map indicate areas where surfaces are present, contrasting with gray regions that represent empty spaces. 
During training at 1K and 3K iterations, regions with high SBV IG in intra-image normalization consistently align with areas where surfaces are present. 
Remarkably, even in early training stages where surface inference is ambiguous and normal maps lack clarity, SBV effectively identifies regions of high color uncertainty through surface-based visibility.
An analysis of SBV IG using global normalization reveals a gradual decline in IG magnitude as additional views are incorporated and training progresses. 
This reduction in IG across test views serves as empirical evidence of effective model convergence.
However, the SBV IG visualization, based on sampling from the Nerfacc \cite{li2023nerfacc} grid, also exposes persistent background noise patterns that remain unaffected by reconstruction quality. 
This artifact highlights an area for further investigation in future studies.

\begin{table*}[!t]  
  \caption{Evaluation of image rendering on DTU, Blender, TanksAndTemples, and BlendedMVS. 
  }\label{tab:render_all}
  \centering
  \begin{adjustbox}{width=1.0\linewidth,center}
  \begin{tabular}{l c c c c c c c c c c c c}
   &\multicolumn{3}{c}{DTU}&\multicolumn{3}{c}{Blender}&\multicolumn{3}{c}{TanksAndTemples}&\multicolumn{3}{c}{BlendedMVS}\\
   \cmidrule(lr){2-4}
   \cmidrule(lr){5-7}
   \cmidrule(lr){8-10}
   \cmidrule(lr){11-13}
Methods&PSNR$\uparrow$&SSIM$\uparrow$&LPIPS$\downarrow$&PSNR$\uparrow$&SSIM$\uparrow$&LPIPS$\downarrow$&PSNR$\uparrow$&SSIM$\uparrow$&LPIPS$\downarrow$&PSNR$\uparrow$&SSIM$\uparrow$&LPIPS$\downarrow$\\
\midrule
Random
&\cellcolor{tabthird}27.69&\cellcolor{tabsecond}0.864&\cellcolor{tabsecond}0.170
&16.24&0.831&0.268
&17.17&0.762&0.361
&\cellcolor{tabthird}26.21&\cellcolor{tabfirst}0.887&\cellcolor{tabthird}0.131\\
FVS
&27.09&0.852&0.187
&\cellcolor{tabthird}20.07&\cellcolor{tabfirst}0.873&\cellcolor{tabthird}0.187
&17.94&0.773&0.339
&25.49&0.881&0.140\\
Entropy
&24.21&0.810&0.218
&15.41&0.818&0.291
&16.70&0.731&0.380
&25.72&0.884&0.137\\
ActiveNeRF*
&26.30&0.852&\cellcolor{tabthird}0.175
&19.25&0.841&0.208
&\cellcolor{tabsecond}18.62&\cellcolor{tabsecond}0.791&\cellcolor{tabsecond}0.304
&\cellcolor{tabsecond}26.57&\cellcolor{tabthird}0.885&\cellcolor{tabfirst}0.128\\
FisherRF*
&\cellcolor{tabsecond}27.78&\cellcolor{tabthird}0.860&0.180
&\cellcolor{tabsecond}20.48&\cellcolor{tabsecond}0.857&\cellcolor{tabsecond}0.183
&\cellcolor{tabthird}18.44&\cellcolor{tabthird}0.778&\cellcolor{tabthird}0.333
&25.71&0.883&0.141\\
Ours
&\cellcolor{tabfirst}28.19&\cellcolor{tabfirst}0.867&\cellcolor{tabfirst}0.168
&\cellcolor{tabfirst}21.22&\cellcolor{tabthird}0.854&\cellcolor{tabfirst}0.170
&\cellcolor{tabfirst}20.49&\cellcolor{tabfirst}0.812&\cellcolor{tabfirst}0.273
&\cellcolor{tabfirst}26.80&\cellcolor{tabsecond}0.886&\cellcolor{tabsecond}0.130\\
  \end{tabular}
  \end{adjustbox}
\end{table*}

\textbf{Performance Comparison.} Our method outperformed other NBV selection methods in both image rendering and mesh reconstruction, as demonstrated quantitatively (\cref{tab:render_all}, \cref{tab:mesh_dtu} in the supplementary) and qualitatively.
 In \cref{tab:render_all} and \cref{tab:mesh_dtu}, the best, second-best, and third-best performing methods are highlighted in red, orange, and yellow, respectively. 
 
 Based on the image rendering evaluation criteria (\cref{tab:render_all}), our approach outperformed all other methods across the four benchmarking datasets.
 The second- and third-best performing methods, however, varied across the datasets, indicating that each dataset presents unique complexities and challenges.
 This robust performance across varying datasets highlights the effectiveness of the surface-based visibility-guided uncertainty. A detailed analysis and qualitative results for each dataset are provided in the supplementary material.

\begin{table}[!t]
\caption{Ablation studies on SBV-based view selection using DTU dataset.  
  }
  \centering
  \begin{adjustbox}{width=1.0\linewidth,center}
    \begin{tabular}{c c c c c c c}
Surface&PSNR$\uparrow$&SSIM$\uparrow$&LPIPS$\downarrow$&Acc.$\downarrow$&Comp.$\downarrow$&Chamfer$\downarrow$\\
\cmidrule(lr){1-1}
 \cmidrule(lr){2-7}
Y&\textbf{28.19}&\textbf{0.867}&\textbf{0.168}&\textbf{1.829}&\textbf{2.176}&\textbf{2.002}\\
N&27.18&0.855&0.180&2.060&2.441&2.251\\
  \end{tabular}
  \end{adjustbox}
  \label{tab:ablation}
\end{table}

\subsection{Ablation Studies}
We analyze the effects of surface-based visibility field in image rendering and mesh reconstruction performance. 
 The comparison method that does not consider SBV estimates uncertainties by integrating them over all ray-traversed voxels, $\forall x \in \mathcal{X}_r$ in $G_s(v)$. 
In \cref{tab:ablation}, SBV-based view selection consistently outperformed the other in all evaluation criteria.
The compact camera view distribution in DTU allowed sufficient coverage of target objects with fewer views, emphasizing the importance of selecting informative views that can enhance incomplete reconstruction.
SBV-based view selection can better focus on selecting informative views than the comparison method that does not consider SBV.
Further qualitative results and analysis are provided in the supplementary material.

\subsection{New Dataset with Imbalanced Viewpoints}
 \label{new_dataset_main}
\begin{figure}[!t]
\centering
     \includegraphics[width=1.0\linewidth]{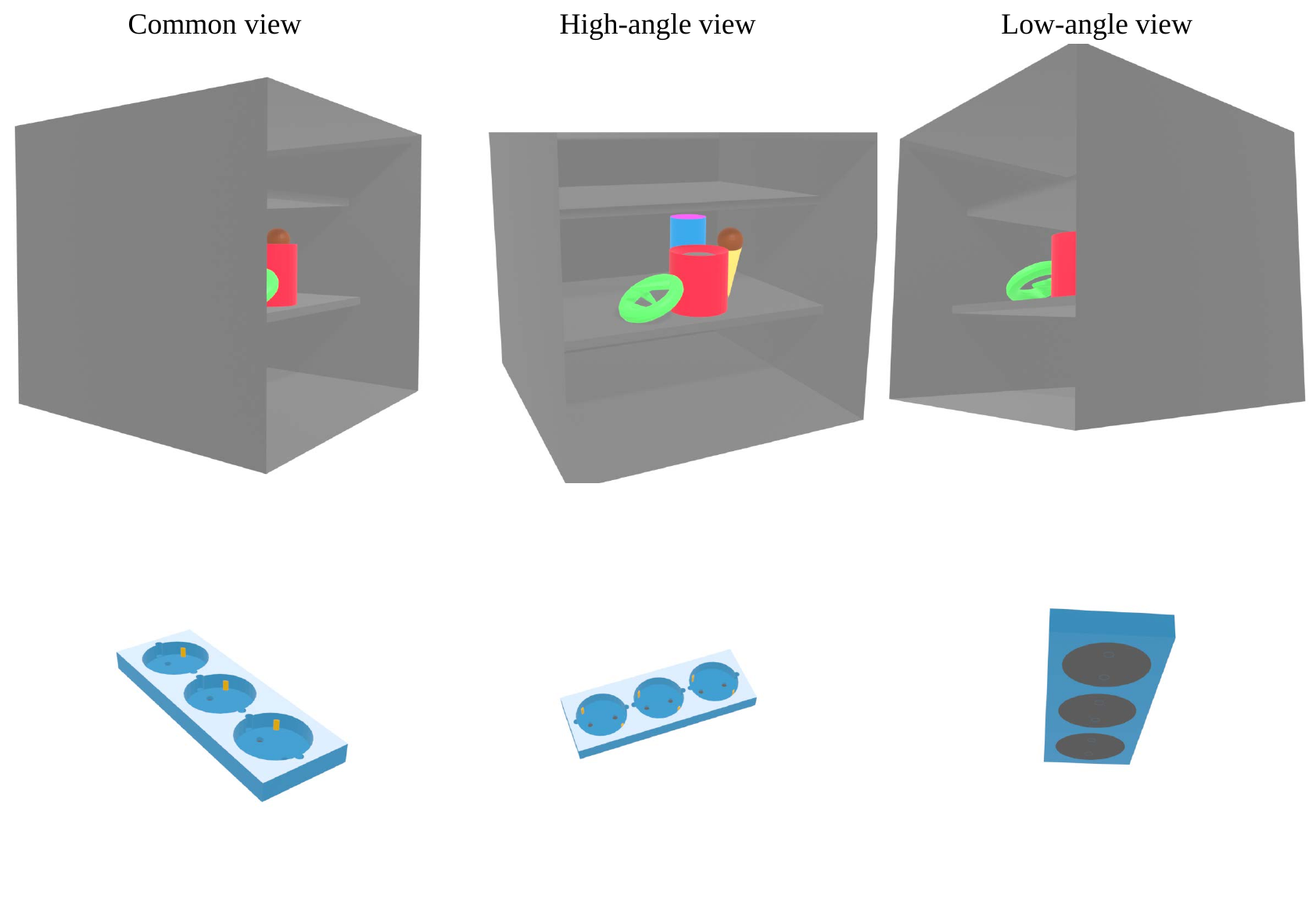}
    \caption{
    Types of viewpoints in the ImBView dataset. Distribution of view types: common view (75\% of the training set, 40\% of the test set), high-angle view (12.5\% of the training set, 40\% of the test set), low-angle view (12.5\% of the training set, 20\% of the test set).
     }
     \label{fig:custom_dataExplain}
 \end{figure}

 We developed a new 3D reconstruction dataset, ImBView, designed to capture occluded objects with imbalanced viewpoints.  
 This dataset is made for the visual analysis of view selection strategies and the intentional restriction of information that randomized view selection can get.
The dataset comprises 80 training images and 10 test images across two scenes: the shelf and the outlet. \cref{fig:custom_dataExplain} illustrates the view types in the dataset. The training set includes 60 common views, 10 high-angle views, and 10 low-angle views, while the test set comprises four common views, four high-angle views, and two low-angle views.
 The dataset's imbalanced nature provides unique perspectives that highlight its practical relevance. 
 For example, the upper side of the blue cylinder within the shelf is only visible in high-angle views. Similarly, in the outlet scene, both black holes can be observed simultaneously only in high-angle views; in common views, only one black hole is visible at a time.
 To reconstruct target objects without missing any specific properties, the model should select views with a balance in types for successful model convergence.

 \begin{table}
\caption{PSNR comparison on the ImBView dataset
}
  \begin{adjustbox}{width=1.0\linewidth,center}
\begin{tabular}{l cccccc}
Scene & Random & FVS & Entropy & ActiveNeRF* & FisherRF* & Ours \\
\hline
Shelf & 22.99 & 17.86 & 15.99 & \cellcolor{tabsecond}25.31 & \cellcolor{tabthird}23.29 & \cellcolor{tabfirst}28.97 \\
Outlet & \cellcolor{tabthird}32.77 & 32.40 & 32.31 & \cellcolor{tabsecond}33.89 & 32.25 & \cellcolor{tabfirst}35.49 \\
\hline
Mean & \cellcolor{tabthird}27.88 & 25.13 & 24.15 & \cellcolor{tabsecond}29.6 & 27.77 & \cellcolor{tabfirst}32.23 \\
\end{tabular}
  \end{adjustbox}
  \label{tab:psnr_custom}
\end{table}

\begin{figure}[!t]
     \centering
\includegraphics[width=1.\linewidth]{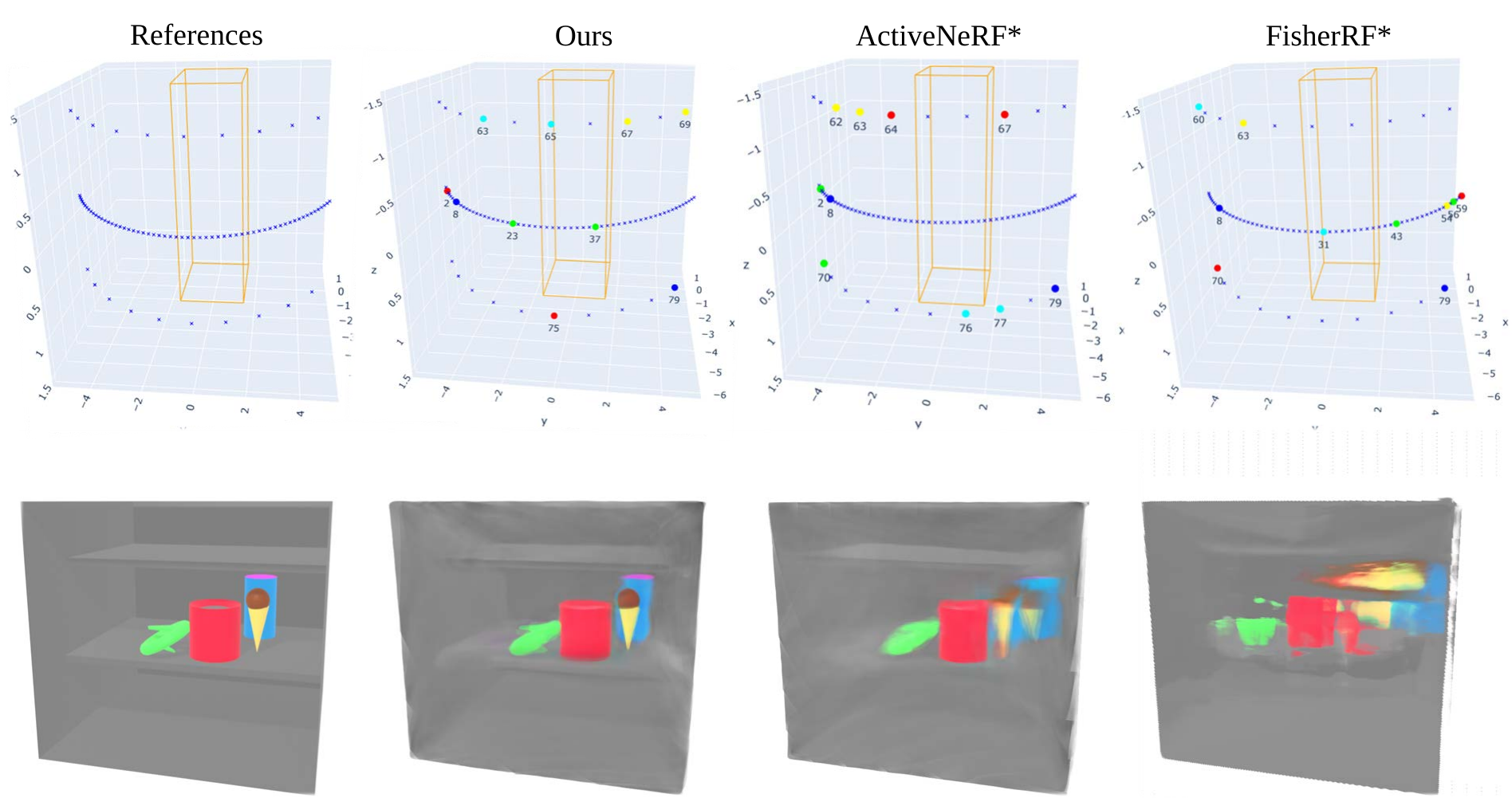}
     \caption{
     View selections and rendered images of three best-performing NBV selections using ImBView dataset in two-image setting are compared. 
     Results of the view selection are shown in the order of blue-cyan-green-yellow-red.
     The two initial views (shown in blue) are fixed in all NBV selection methods.
     The test image is from a high-angle view.
}
\label{fig:custom_shelf}
 \end{figure}

We compared the quality of image renderings produced by different NBV methods using the ImBView dataset in a two-image setting (\cref{tab:psnr_custom}).
As depicted in \cref{fig:custom_shelf}, FisherRF* exhibited a bias toward selecting views at the right and left ends. 
Example views from these ends are illustrated as the common view (left) and low-angle view (right) in~\cref{fig:custom_dataExplain}.
The central views, represented as high-angle view in \cref{fig:custom_dataExplain}, provide visibility of all four objects and their colors.
However, FisherRF* predominantly selected views from the ends, resulting in low-informative views and poor image rendering outcomes.

ActiveNeRF* selected central views more frequently than FisherRF* but still failed to accurately capture the 3D structure. For instance, it misplaced the chocolate ice cream on the shelf.
In contrast, our method effectively selected views from diverse angles, successfully reconstructing all four objects with accurate colors.
 This highlights the robustness of our approach in addressing imbalanced viewpoints. Additional results regarding our new dataset can be found in the supplementary material.

\subsection{Discussions and Limitations}

We highlight two limitations and associated discussions of our approach: 
\textit{i}): Our present work does not consider displacement costs between views, which is critical for comprehensive view planning in robotics \cite{ran2023neurar, jin2023neunbv, zhan2022activermap, feng2024naruto, yan2023active}.
\textit{ii}): Our neural implicit surface network relies on frequency encoding rather than multi-resolution hash encoding methods such as NeuS2 \cite{neus2} and Neuralangelo \cite{li2023neuralangelo}.

Our approach is not designed for reconstructing indoor scenes, which often require depth information \cite{Yu2022MonoSDF}.
Consequently, the displacement costs between views, a key component of view planning for indoor scenes, are not considered. 
Future research could extend SBV to accommodate indoor environments using depth data.  

Lastly, we avoided multi-resolution hash encoding due to its locality issues when calculating gradients for Eikonal loss~\cite{li2023neuralangelo}.
Our NBV selection method requires accurate computation of surface confidence for SBV-guided IG computation. 
The locality problem inherent to hash encoding could result in inaccurate surface estimation, potentially degrading view selection quality.
Although Neuralangelo~\cite{li2023neuralangelo} mitigates these limitations using numerical gradients, the increased GPU memory usage and computational cost provide no substantial advantages over NeuS for our objectives. 

Our NBV selection method remains adaptable to any network learning ray-based surface and rendering uncertainty.
These discussions and limitations highlight critical areas for future exploration and development.

\section{Conclusion}
We have presented SBV, a surface-based visibility field developed for measuring uncertainty in the NBV selection framework. 
SBV effectively identifies both well-defined surfaces and ambiguous regions in scene geometry through surface confidence and computes accurate IG, thereby significantly improving 3D reconstruction performance in complex and occluded scenes compared to existing methods.
For future work, applying SBV-based view selection to robotic active 3D reconstruction, where a robot arm dynamically moves and collects data, would be an exciting direction to pursue.  

\bibliography{main}

\appendix
\clearpage

\section{Experimental Details and Results}
\label{sec:appendix_results}

\subsection{Comparison of Methods}
\label{sec:appendix_methodcomp}

We elaborate on the details of the next-best view (NBV) selection methods: ActiveNeRF, Entropy, FisherRF, and FVS.
Note that all the methods are implemented based on SDFStudio \cite{Yu2022SDFStudio}.
We used the code in SDFStudio that implements grid sampling, which is similar to the accelerated grid sampling utilized by NerfAcc \cite{li2023nerfacc} and Instant NGP \cite{muller2022instant}.
The grid sampling method selects 3D points by referencing an occupancy grid. During ray marching, the grid-sampled points with an occupancy probability lower than a given threshold are skipped. 
Therefore, with grid sampling, a variable number of samples are selected in each ray. 
We employed this grid sampling technique for efficient and rapid training.

In ActiveNeRF, we employed a different sampling method for information gain (IG) calculation, which samples a constant number of points along a ray as in the original paper \cite{pan2022activenerf}, instead of the grid sampling used during training.

For the entropy method with surface representation, we substitute $\alpha$ defined in NeuS (Eq. (2)) for the occupancy probability $p(x)$ used in the entropy calculation (Eq. (6)). As $p(x)$ is the occupancy probability of a voxel, its value is in the range of 0 to 1, and the initial value of $p(x)$ is 0.5 to make the entropy calculated from it to be the highest in the initialization.
Similarly, $\alpha$ is 0 when the unit opaque density $\rho$ shows a low value and 1 when the unit opaque density $\rho$ shows a high value.

For FisherRF*, we changed the base model architecture from 3D Gaussians \cite{kerbl3Dgaussians} to NeuS \cite{wang2021neus}.
This change in architecture resulted in an increase in the overall time required for training and view selection.
Two factors contribute to this increased time consumption: (1) differences in CUDA implementation between 3D Gaussians and NeuS, and (2) the increased number of model parameters for Fisher information calculation.
Our method requires less time for view selection but more time for overall training. 
If the focus is on reducing overall time consumption rather than memory requirements for training, 3D Gaussian Splatting methods that reconstruct surfaces can be used~\cite{guedon2023sugar, chen2023neusg, Huang2DGS2024, Dai2024GaussianSurfels, Wu2024gsrec, zhang2024gspull}.
However, they have some limitations, such as requiring surface information estimated from the neural implicit surface networks \cite{chen2023neusg} or requiring additional information such as normal priors or depth information \cite{Dai2024GaussianSurfels, Wu2024gsrec}.
Additionally, there may be a trade-off between image rendering quality and mesh reconstruction quality in some cases~\cite{guedon2023sugar, Huang2DGS2024, zhang2024gspull}; thus, the choice of models should be based on specific use cases.

\begin{table}
\caption{Details of network architectures 
}
\begin{center}
\resizebox{\columnwidth}{!}{%
\begin{tabular}{clccc}
{} & \textbf{Networks} & {Surface} & {Density} & {ActiveNeRF~\cite{pan2022activenerf}} \\
\midrule
\textbf{Density}&MLP hidden layer & 8 & 8 & 8 \\
\textbf{or}&MLP size & 256 & 256 & 256 \\
\textbf{SDF}&Activation & Softplus & ReLU & ReLU  \\
\textbf{network}&Positional encoding & 6 & 10 & 10  \\
&Skip connection layer & 4 & 4 & 4  \\
\midrule
\textbf{Color}&MLP hidden layer & 4 & 1 & 1  \\
\textbf{network}&MLP size & 256 & 128 & 128  \\
&Direction encoding & 4 & 4 & 4  \\
\midrule
\textbf{Hyper-}&RGB loss & 1 &1 & 1  \\
\textbf{parameters}&Eikonal loss & 0.1 & - &- \\
&Batch size & 512 & 512 & 1024  \\
&Learning rate & $5 \times 10^{-4}$ & $5 \times 10^{-4}$ & $5 \times 10^{-4}$ \\
\midrule
\textbf{Uncertainty-}&Uncertain loss (RGB) & 0.001 & 0.001 & 1  \\
\textbf{related}&Uncertain loss ($\beta$) & 0.01 & 0.01 & 0.5  \\
\textbf{hyper-}&Uncertain loss ($\sigma$) & 0.0 & 0.0 & 0.01  \\
\textbf{parameters}&Minimum of $\beta$ & 0.001 & 0.001 & 0.01 
 \\
\bottomrule
\end{tabular}
}
\label{hyperparameter}
\end{center}
\end{table}

Lastly, for FVS, we used FVS(euc) in NeRFDirector \cite{Xiao:CVPR24:NeRFDirector}. 
NeRFDirector suggested three method types (two for spatial distance, one for photogrammetric distance) for measuring distances for FVS. However, we opted for the Euclidean distance, which is a commonly used method for measuring distances in FVS.
Additionally, the authors reported minimal performance difference between the FVS using both spatial and photogrammetric distances and FVS using the Euclidean distance only.

We further compared our SBV-based view selection strategy with other NBV selection methods implemented within the NeRF framework. 
This comparison was necessary because the original IG formulation of the entropy method and ActiveNeRF are based on density representation, not surface representation.

Before we demonstrate the quantitative and qualitative comparison results, first, we elaborate on the architectural differences among three types of models: neural networks encoding surface representations, those encoding density representations, and ActiveNeRF \cite{pan2022activenerf}.
In \cref{hyperparameter}, we used the NeuS \cite{wang2021neus} architecture for models with surface representation and the NeRF \cite{mildenhall2021nerf} architecture for networks with density representation. 
Except for ActiveNeRF~\cite{pan2022activenerf}, all networks are implemented based on SDFStudio \cite{Yu2022SDFStudio}.
The uncertainty-related hyperparameters are utilized when the model employs ActiveNeRF- or SBV-based view selection. 
The coefficients of these hyperparameters differ from those in the original implementation of ActiveNeRF \cite{pan2022activenerf}. 
Specifically, we adjusted the coefficients to stabilize model training during the early stages, when only a small number of training images are available.
The impact of these modifications is explained using the BlendedMVS dataset.

 Note that ActiveNeRF and FisherRF implemented in surface representations are named as ActiveNeRF* and FisherRF*. 
The additional results are shown as \cref{tab:select_blender} and \cref{tab:select_dtu}. 

\subsection{Experimental Settings}
\label{appendix:exp_setting}

 \begin{figure}
     \centering
      \includegraphics[width=.9\linewidth]{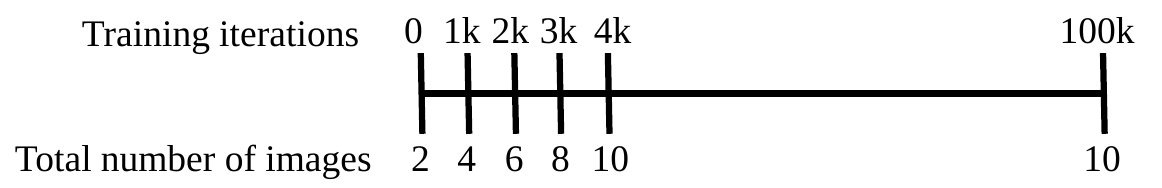}
     \caption{
 Active learning scheme in the two-image setting. 
 Views are selected during the early training stages to evaluate the accuracy in identifying uncertainties from observable regions. 
}
\label{fig:training_scheme}
 \end{figure}

In this subsection, we provide a detailed explanation of the active learning scheme (\cref{fig:training_scheme}), including how initial training image sets are selected.
We then describe minor techniques implemented to stabilize model training during early stages when few training images are available: frequency regularization and warm-up stages.

\textbf{Datasets and metrics.} We evaluated our SBV-based view selection and other NBV selection methods on five datasets using active learning schemes: 15 scenes from the DTU dataset \cite{jensen2014dtu}, eight scenes from the NeRF Blender Synthetic dataset (Blender) \cite{mildenhall2021nerf}, five scenes from the TanksAndTemples dataset \cite{TNT2017}, five scenes from the BlendedMVS dataset \cite{yao2020blendedmvs}, and two scenes from our new dataset, ImBView, which features imbalanced viewpoints. 
We used the Blender and DTU datasets provided by SDFStudio \cite{Yu2022SDFStudio} and the TanksAndTemples dataset from NSVF \cite{liu2020nsvf}. 
From the original 113 scenes in the BlendedMVS dataset, we selected five large-scale scenes with aerial views instead of ground-level perspectives.

In the DTU dataset, each scene comprises 49 or 64 images with a resolution of $384\times384$, with 10 images reserved for testing. 
For the Blender dataset, each scene includes 100 training images and 200 test images at a resolution of $800\times800$; however, we used 25 evenly sampled images for testing.
In TanksAndTemples dataset, each scene contains 152 to 384 images with a resolution of $960\times540$, with 1/8 of the images reserved for testing. 
For BlendedMVS dataset, each scene contains 77 to 339 images with a resolution of $768\times576$, with 1/8 of the images reserved for testing.
The ImBView dataset includes 80 training images and 10 test images, each with a resolution of $800\times800$.

\begin{figure}[!t]
      \centering
\includegraphics[width=0.9\linewidth]{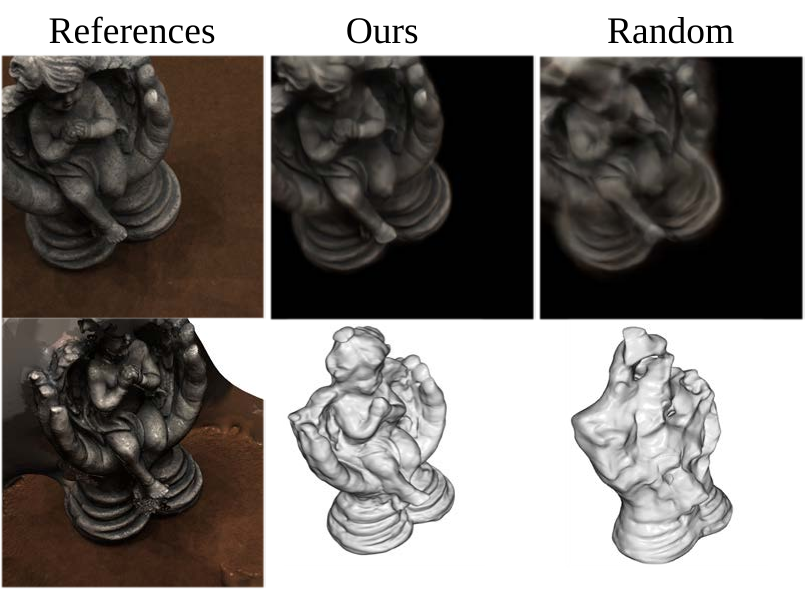}
     \caption{
     Rendered images and reconstructed meshes of two best-performing algorithms (in both criteria: mesh reconstruction and image rendering) using DTU dataset in two-image setting. 
     Second row of references shows ground truth point cloud instead of mesh.
}
\label{fig:dtu_rgb}
 \end{figure}

The training images are divided into a training set and a candidate set, where the candidate set consists of all training images except those included in the training set. During candidate view evaluation for NBV selection, only the camera pose matrices from the candidate set are accessible; ground-truth images become accessible only after the selection.

 For the DTU and BlendedMVS datasets, all NBV methods were trained with object masks applied to the images.
 NeuS-based methods typically require an additional model (\eg, NeRF) for background training \cite{wang2021neus}. 
In this study, we focused exclusively on the target object by applying object masks, thereby avoiding the need to address the integration of uncertainties from different network types. 
The DTU dataset provided by SDFStudio includes object masks, while for the BlendedMVS dataset, object masks were generated using SAM2 \cite{ravi2024sam2}. 

Five scenes were selected for training where SAM2 exhibited consistent and reliable object segmentation masks. 
These scenes primarily feature views that comprehensively capture the shapes of the target objects, allowing SAM2 to effectively distinguish targets from the background. 
Scenes with ambiguous boundary delineations were excluded, as their segmentation accuracy could not be reliably verified.

Image rendering quality was evaluated using peak signal-to-noise ratio (PSNR), structural similarity index measure (SSIM) \cite{SSIM}, and LPIPS \cite{zhang2018lpips} scores, following the metrics established by ActiveNeRF \cite{pan2022activenerf}. 
The reconstructed meshes were evaluated using three metrics-accuracy, completeness, and Chamfer distance-consistent with the evaluation methodology in ActiveRMAP \cite{zhan2022activermap}. 

 \begin{figure}[!t]
     \centering
\includegraphics[width=1.\linewidth]{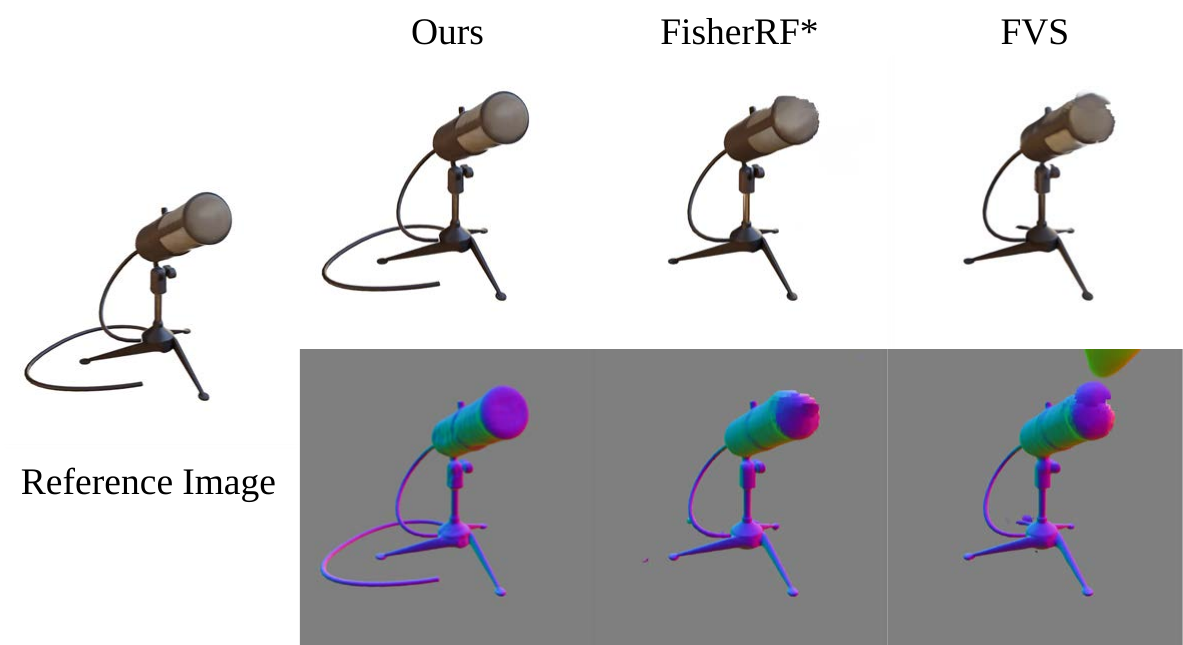}
     \caption{
     Rendered novel views and corresponding normal maps of three best-performing NBV selection methods using Blender dataset in two-image setting.
     First and second rows show rendered novel views and corresponding normal maps, respectively.     
}
\label{fig:blender_mic}
 \end{figure}

  \begin{figure}[!t]
     \centering
\includegraphics[width=1.\linewidth]{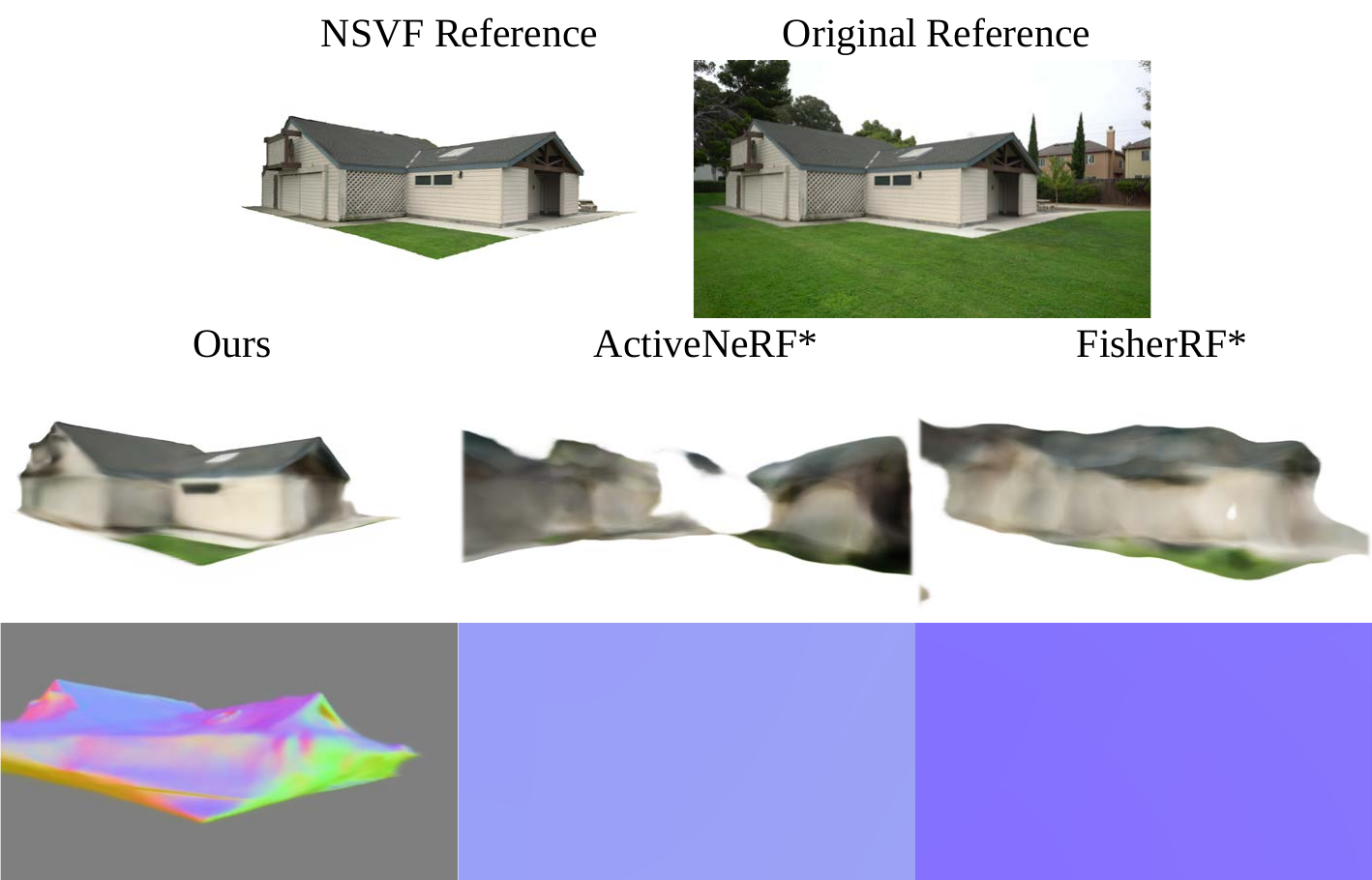}
     \caption{
     Rendered novel views and corresponding normal maps of three best-performing NBV selection methods using TanksAndTemples dataset in ten-image setting. We used TanksAndTemples dataset from NSVF \cite{liu2020nsvf}.
     First, second, and third rows show reference images of barn, rendered novel views, and corresponding normal maps, respectively.
}
\label{fig:tnt}
 \end{figure}

\textbf{Experiment protocol.} The total training iterations in the Blender dataset's one-, two-, and four-image settings are $50k$, $100k$, and $200k$, respectively. 
For selection intervals, in the one-image setting, the model selects the NBV every $0.5k$ iterations (\eg, $[0.5k, 1k, 1.5k, 2k]$). In the two-image setting, NBV selection occurs every $1k$ iterations (\eg, $[1k, 2k, 3k, 4k]$). 
Similarly, in the four-image setting, NBV selection occurs every $2k$ iterations (\eg, $[2k, 4k, 6k, 8k]$).
As the model selects NBVs four times, the total numbers of training images are 5, 10, and 20 in one-, two-, and four-image settings, respectively.
The initial training image set of Blender before the view selection is $[26, 86, 2, 55, 75, 93, 16, 73, 8]$, following the experimental setting in FreeNeRF \cite{yang2023freenerf}. For example, in the two-image setting, the initial training images are $[26, 86]$, and in the four-image setting, the initial training images are $[26, 86, 2, 55]$. 

For the ImBView dataset, we trained models for $100k$ iterations in a two-image setting, and the views are selected every $1k$ iterations in the two-image setting (\eg, $[1k, 2k, 3k, 4k]$). The initial training images are $[8, 79, 68, 12]$ (which means $[8, 79]$ are used in a two-image setting).
For the Blender and ImBView datasets, which feature white backgrounds and centrally positioned target objects, we applied a center-cropping strategy with varying cropping ratios and durations depending on the number of training images. 
In the one-image setting, images were center-cropped to $0.75\times$ until 400 training iterations. For the two-image and four-image settings, images were center-cropped to $0.6\times$, with durations of 750 and 1200 training iterations, respectively.

For the DTU dataset, we trained models for $60k$ iterations in a two-image setting and $120k$ iterations in a four-image setting. The selection intervals are the same as the Blender dataset: every $1k$ iterations in the two-image setting (\eg, $[1k, 2k, 3k, 4k]$), and every $2k$ iterations in the four-image setting (\eg, $[2k, 4k, 6k, 8k]$).
The initial training images in DTU are $[25, 22, 28, 40, 44, 48, 0, 8, 13]$, following FreeNeRF. Additionally, the test image set in DTU is $[1, 9, 12, 15, 24, 27, 32, 35, 42, 46]$.

For the TanksAndTemples and BlendedMVS datasets, the total training iterations are $300k$, and the selection intervals are every $5k$ iterations in a ten-image setting (\eg, $[5k, 10k, 15k, 20k]$). 
During training in the BlendedMVS dataset, we used different coefficients for uncertain losses (\cref{hyperparameter}). The uncertain loss (RGB) coefficient decreased from 0.001 to 0.0007, and the uncertain loss (beta) coefficient increased from 0.01 to 0.04. 
Empirically, we observed that an excessively high uncertain loss (RGB) coefficient negatively impacts the training gradient force from the original color loss.
This effect can impede color-based 3D reconstruction training, particularly in the early stages.
Consequently, for the complex scenes in BlendedMVS with high color diversity, we reduced the uncertain loss (RGB) coefficient to ensure stable model training. Instead, we increased the uncertain loss (beta) coefficient to mitigate color variance.

Since no initial training sets or test sets were reserved for the active 3D reconstruction of the TanksAndTemples and BlendedMVS published, we heuristically select the training and test sets that cover the whole 360-degree view. Note that the image indexes are numbered after the datasets are post-processed in SDFStudio format. 

TanksAndTemples dataset:
\begin{itemize}
    \item Truck (initial training set): \\
    $[16, 18, 19, 20, 23, 26, 141, 32, 207, 132]$
    \item Truck (test set): \\ 
    $[121,  92,  49, 184, 165,  89, 123, 193, 212,  34,  66, \\
    175, 144, 138,  90, 210, 179, 111, 185, 177, 146,\\ 107,  71, 125,  36, 233, 170, 215, 136, 130,  31, 217]$
    \item Barn (initial training set): \\ $[8, 20, 17, 16, 19, 26, 32, 100, 67, 120]$
    \item Barn (test set): \\
    $[206, 205, 259, 325, 236, 185,  46, 119, 194, 253,\\  61, 183, 281, 304, 225, 180, 293,  96,  55, 177,\\ 241, 202, 358, 27, 243, 251,  39, 112, 272,  12, \\ 190,  34, 214,  30, 292, 254, 165, 174, 328, 296, \\ 179, 321, 246, 198, 258, 306, 140, 343]$
    \item Ignatius (initial training set): \\ $[30, 65, 75, 38, 76, 119, 154, 172, 173, 174]$
    \item Ignatius (test set): \\
    $[256,  86, 237,  20, 258, 121,  24,  84, 122,  66, 240,\\  64, 199, 42, 202,  37, 198, 254, 159,  46, 139, \\ 53,  89,  32, 230,  96,  98, 135, 227, 197, 241, 247, 163]$
 \item Family (initial training set): \\
 $[50, 47, 19, 12, 43, 92, 90, 136, 148, 130]$
 \item Family (test set): \\
 $[ 48,  18, 118,  77, 142,  27, 101,  98, 107,   6,\\ 109,  42,  15, 124, 104,  93, 129, 102,  13]$
 \item Caterpillar (initial training set): \\
 $[5, 54, 104, 117, 148, 135, 55, 50, 124, 89]$
 \item Caterpillar (test set): \\
 $[189,  40, 244,  42,  26, 255,  53,  63, 357, 197,\\  10,  18, 287, 72, 303,  57, 134, 164, 329, 202, 246, \\ 320,  45, 300,  86, 301,  200, 167, 245, 144, 241,  84, \\ 239, 235, 224, 261, 161,  58, 298,  271,  35, 141, \\ 313, 166,  68, 342]$
\end{itemize}

\begin{table*}[!t]
  \centering
  \caption{
  Evaluation of image rendering on Blender. 
  }
  \resizebox{1.0\linewidth}{!}{%
  \begin{tabular}{l c c c c c c c c c }
   &\multicolumn{3}{c}{one-image setting}&\multicolumn{3}{c}{two-image setting}&\multicolumn{3}{c}{four-image setting}\\
   \cmidrule(lr){2-4}
   \cmidrule(lr){5-7}
   \cmidrule(lr){8-10}
Methods&PSNR$\uparrow$&SSIM$\uparrow$&LPIPS$\downarrow$&PSNR$\uparrow$&SSIM$\uparrow$&LPIPS$\downarrow$&PSNR$\uparrow$&SSIM$\uparrow$&LPIPS$\downarrow$\\
 \midrule
\textbf{Density (except Ours)} \\
\midrule
Random&11.05&0.744&0.442&12.34&0.759&0.404&\cellcolor{tabthird}26.68&\cellcolor{tabthird}0.909&\cellcolor{tabthird}0.089\\
FVS&11.08&0.738&0.443&12.38&0.760&0.417&\cellcolor{tabfirst}27.07&\cellcolor{tabfirst}0.916&\cellcolor{tabfirst}0.085\\
Entropy&11.45&\cellcolor{tabthird}0.752&0.423&12.70&0.779&0.383&25.71&0.899&0.100\\
ActiveNeRF &\cellcolor{tabsecond}13.09&\cellcolor{tabsecond}0.776&\cellcolor{tabsecond}0.362&\cellcolor{tabthird}16.64&\cellcolor{tabthird}0.807&0.270&18.82&0.822&0.261\\
ActiveNeRF~\cite{pan2022activenerf}&-&-&-&\cellcolor{tabsecond}20.01&\cellcolor{tabsecond}0.832&\cellcolor{tabsecond}0.204&26.24&0.856&0.124\\
ActiveNeRF (repro.)&-&-&-&14.78&0.792&\cellcolor{tabthird}0.251&17.00&0.820&0.200\\
FisherRF&\cellcolor{tabthird}11.51&0.752&\cellcolor{tabthird}0.420&12.99&0.765&0.397&\cellcolor{tabsecond}26.82&\cellcolor{tabsecond}0.913&\cellcolor{tabsecond}0.087\\
Ours&\cellcolor{tabfirst}16.63&\cellcolor{tabfirst}0.809&\cellcolor{tabfirst}0.265& \cellcolor{tabfirst} 21.22& \cellcolor{tabfirst} 0.854& \cellcolor{tabfirst} 0.170&26.08&0.907&0.106\\
\midrule
\textbf{Surface} \\
\midrule
Random&15.88&\cellcolor{tabthird}0.812&\cellcolor{tabthird}0.268&16.24&0.831&0.268&\cellcolor{tabthird}21.79&0.874&\cellcolor{tabthird}0.151\\
FVS&\cellcolor{tabfirst}16.64&\cellcolor{tabfirst}0.813&\cellcolor{tabfirst}0.258&\cellcolor{tabthird}20.07&\cellcolor{tabfirst}0.873&\cellcolor{tabthird}0.187&21.57&\cellcolor{tabthird}0.876&0.154\\
Entropy&13.78&0.801&0.307&15.41&0.818&0.291&21.28&0.869&0.159\\
ActiveNeRF* &\cellcolor{tabthird}15.93&0.809&0.270&19.25&0.841&0.208&\cellcolor{tabsecond}23.91&\cellcolor{tabsecond}0.884&\cellcolor{tabsecond}0.127\\
FisherRF*&15.83&\cellcolor{tabsecond}0.813&0.279&\cellcolor{tabsecond}20.48&\cellcolor{tabsecond}0.857&\cellcolor{tabsecond}0.183&20.66&0.867&0.172\\
Ours&\cellcolor{tabsecond}16.63&0.809&\cellcolor{tabsecond}0.265& \cellcolor{tabfirst} 21.22& \cellcolor{tabthird} 0.854& \cellcolor{tabfirst} 0.170&\cellcolor{tabfirst}26.08&\cellcolor{tabfirst}0.907&\cellcolor{tabfirst}0.106 \\
  \bottomrule
  \end{tabular}
  }
  \begin{minipage}{\textwidth}
  \vspace{0.1cm}
      \small
In the one-image setting with density representation, the entropy method is highlighted as the third-best performing method due to its marginally higher value (0.7519625 versus 0.7515125 for FisherRF*). 
Similarly, in the one-image setting with surface representation, FVS is designated as the best-performing method owing to its slightly higher value (0.8133625 compared to 0.8127125 for FisherRF*). Furthermore, we report both the original results from the ActiveNeRF \cite{pan2022activenerf} paper and our reproduced results, ActiveNeRF(repro.), using the official ActiveNeRF implementation.
  \end{minipage}
   \label{tab:select_blender}
\end{table*}

BlendedMVS dataset:
\begin{itemize}
    \item man (initial training set): \\$[0, 7, 11, 89, 56, 57, 83, 81, 31, 27]$
    \item man (test set):\\ $[102, 91, 86, 77, 66, 52, 43, 35, 24, 4, 
                        40, 53, 74]$
    \item cathedral (initial training set): \\$[13, 77, 61, 52, 41, 74, 80, 87, 96, 100]$
    \item cathedral (test set):\\ $[71, 65, 59, 55, 8, 75, 81, 84, 95, 103,
                        117, 69, 114, 90, 76]$
    \item dragon (initial training set): \\$[0, 9, 15, 24, 28, 52, 64, 75, 110, 121]$
    \item dragon (test set):\\ $[7, 4, 11, 17, 8, 6, 23, 30, 39, 43, 
       \\48, 55, 71, 93, 109, 117, 126, 136, 147, 152,
        \\ 173, 176, 185, 193, 204, 211, 218, 231, 241, 250,
    \\ 337, 325, 316, 309, 298, 293, 284, 277, 274, 264, 247, 84]$
    \item redDome (initial training set): \\$[0, 2, 6, 45, 14, 56, 66, 18, 29, 43]$
    \item redDome (test set): \\$[74, 69, 62, 52, 48, 49, 21, 20, 24]$
    \item pavilion (initial training set): \\$[0, 12, 35, 40, 61, 77, 104, 194, 209, 242]$
    \item pavilion (test set): \\$[269, 290, 272, 187, 217, 199, 151, 7, 16, 27, 
 \\  36, 45, 55, 68, 78, 89, 95, 106, 113, 125,
 \\   136, 148, 167, 178, 205,  223, 234, 244, 256, 285,
  \\  262, 18, 28, 102, 163, 247,189]$
\end{itemize}

\textbf{Frequency regularization} We applied frequency regularization in the one- and two-image settings, as suggested in FreeNeRF \cite{yang2023freenerf}. 
Given the challenge of learning from few images, frequency regularization helped the model perceive the target with low frequency and gradually transitioned to high frequency as more images were selected and trained. This approach prevented the model from overfitting to a limited number of images.
Following a setup similar to FreeNeRF \cite{yang2023freenerf}, we applied frequency regularization with varying durations across different settings. 
For the two-image setting in Blender, regularization ends at $40k$ iterations (40\% of total iterations). In the one-image setting for Blender, regularization ends at $30k$ iterations (60\% of total iterations). Lastly, when training models with the two-image setting using the DTU dataset, frequency regularization terminates at $30k$ iterations (50\% of total iterations).

\begin{algorithm*}[!t]
\caption{Multiple next-best view selection}
\label{alg:cand_k_view}
\begin{algorithmic}[1]
\State \textbf{Input:} Candidate camera index array $\mI_C$, train camera index array $\mI_T$, camera pose array $\mP$, information gain $G_s$, distance threshold $\tau$, number of views to select $K$. 
\State \textbf{Output:} Next-best view index array $\mI_A$.
\State Select the index $idx$ from $\mI_C$ with a maximum $G_s$ value.
\State Remove $idx$ from $\mI_C$ and add $idx$ to $\mI_A$.
\For{$i \gets 1$ to $K-1$}
    \State Add $idx$ to $\mI_T$ and sort $\mI_C$ in descending order by $G_s$ value.
    \State Compute the pairwise distance matrix $\mD$ between $\mP[\mI_T]$ and $\mP[\mI_C]$.
    \State Get a boolean matrix $\mB = (\mD \geq \tau)$, and perform an "AND" operation along $\mI_T$.      \Comment{The shape of $(\mD \geq \tau)$ is ($\mI_T$, $\mI_C$).}      
    \While{$\mB$ is all False along $\mI_C$.} \Comment{The shape of $\mB$ is ($\mI_C$,).}
        \State $\tau \gets \tau \times 0.95$      \Comment{Lower the distance threshold.}
        \State $\mB = (\mD \geq \tau)$ and do "AND" operation along $\mI_T$          \Comment{Filter again with the relaxed criteria.}
    \EndWhile
    \State $idx = \mI_C[\mB][0]$  \Comment{Select the first element in sorted $\mI_C$ filtered with the distance.}
    \State Remove $idx$ from $\mI_C$ and add $idx$ into $\mI_A$.
\EndFor
\end{algorithmic}
\end{algorithm*}

\textbf{Warm-up} We used two types of warm-up stages, one from the NeuS \cite{wang2021neus} framework and the other from 3D points sampling strategies in NerfAcc \cite{li2023nerfacc}. 
According to NeuS \cite{wang2021neus}, training a model to converge and learn a surface is challenging. To address this, the learning rate of implicit neural surface networks increases linearly in the warm-up stage to stabilize training.
The learning rate of NeuS increases from 0 to $5 \times 10^{-4}$ in the first 500 iterations for the one- and two-image settings, and 1000 iterations for the four-image setting. In a ten-image setting, the warm-up ends at 2500 iterations.

During the warm-up stage in 3D points sampling strategies, a constant number of points along a ray is sampled instead of using grid sampling. 
In addition to sampling a constant number of 3D points, all cells in the NerfAcc occupancy grid are updated. 
By sampling a constant number of 3D points and updating all cells in the occupancy grid, the model can learn information about the whole scene in the early training stages without skipping some parts with grid sampling. 
After the warm-up stage, 3D points are sampled with grid sampling, and the occupancy grid is partially updated by selecting $n_o$ occupied cells and $n_r$ randomly sampled cells, where $n_o$ and $n_r$ are hyperparameters.
In our proposed training scheme, the model reinitiates the warm-up stage for 3D point sampling after each view selection and updates information about the entire scene. 
For example, in the training scheme illustrated as \cref{fig:training_scheme}, the model undergoes 5 warm-up periods in 3D points sampling: $[0,512]$, $[1000, 1512]$, $[2000, 2512]$, $[3000, 3512]$, and $[4000, 4512]$. This approach ensures that the model stably updates the whole scene after the integration of new viewpoints. 
Note that the warm-up stage is lengthened if the model is trained on more images in different settings. 
In one- and two-image settings, the warm-up stage for 3D point sampling persists until the 512th training iteration. For a four-image setting, the warm-up stage extends to the 1024th iteration. In the case of a ten-image setting, the warm-up stage continues up to the 2560th iteration.

 \begin{figure}[!t]
     \centering
\includegraphics[width=1.0\linewidth]{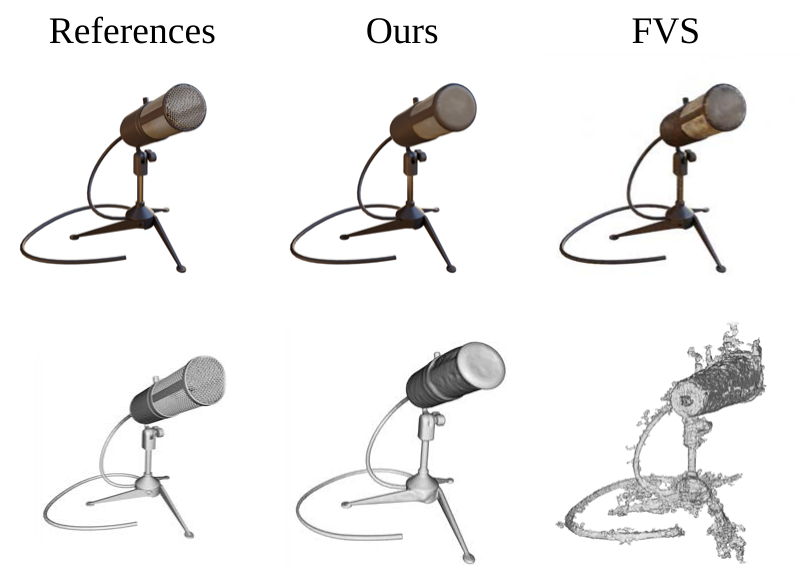}
     \caption{
 Rendered images and reconstructed meshes of the best-performing algorithm and ours in a four-image setting using the Blender dataset are compared. The result from FVS in density representation is visualized.
}
\label{fig:blender_20_supple}
 \end{figure}

\subsection{More Results}
\label{appendix_result}

 For the DTU dataset, we visualized the reconstruction results of scan 106, depicting a baby angel sitting in hands (\cref{fig:dtu_rgb}). The random selection method failed to select views that included incompletely reconstructed areas. 
As a result, it was unable to reconstruct the upper part of the target or the intricate regions inside the hands.
In contrast, our method selected views that captured complex areas, such as the face of the baby angel and the interior of the hands, achieving high accuracy in both reconstruction and rendering.

For the Blender dataset, we visualized the reconstruction results of the microphone scene (\cref{fig:blender_mic}).
The normal maps were visualized in RGB by modifying the value range of the normals with the formula $(\texttt{normal}+1)/2$.
While FisherRF* and FVS failed to reconstruct the challenging thin microphone cord, our method succeeded.
Thin structures are notoriously difficult to reconstruct~\cite{wang2021neus}, often leading to underfitting. 
In such cases, volume rendering can fail to detect the high uncertainty of low-density regions, such as thin cords. 
However, our IG quantification method accurately measured the visibility of these uncertain regions by simultaneously considering surface confidence and uncertainty. Consequently, our method successfully guided the model to reconstruct the thin parts of the microphone.

For the TanksAndTemples dataset, we visualized the RGB rendering and normal estimation results of the top three NBV methods on the barn scene (\cref{fig:tnt}). 
As the reference image from the original dataset shows, the scene captures a large barn from ground-level perspectives instead of aerial views. This setup results in severe occlusions. 
Since the models are trained with only 50 images out of 384 (12.5\%), careful selection of the view is crucial to recognize the entire shape of the barn.
ActiveNeRF* and FisherRF* failed to estimate the normals accurately and struggled to render correct RGB values for novel views. 
This suggests that these methods lacked a comprehensive understanding of the barn's three-dimensional structure, resulting in inconsistent RGB performance across different viewpoints. 
In contrast, our method demonstrated superior performance in both criteria compared to the other two, highlighting the effectiveness of our approach in selecting NBVs by considering surface ambiguity.

 \begin{figure}[!t]
     \centering
\includegraphics[width=1.\linewidth]{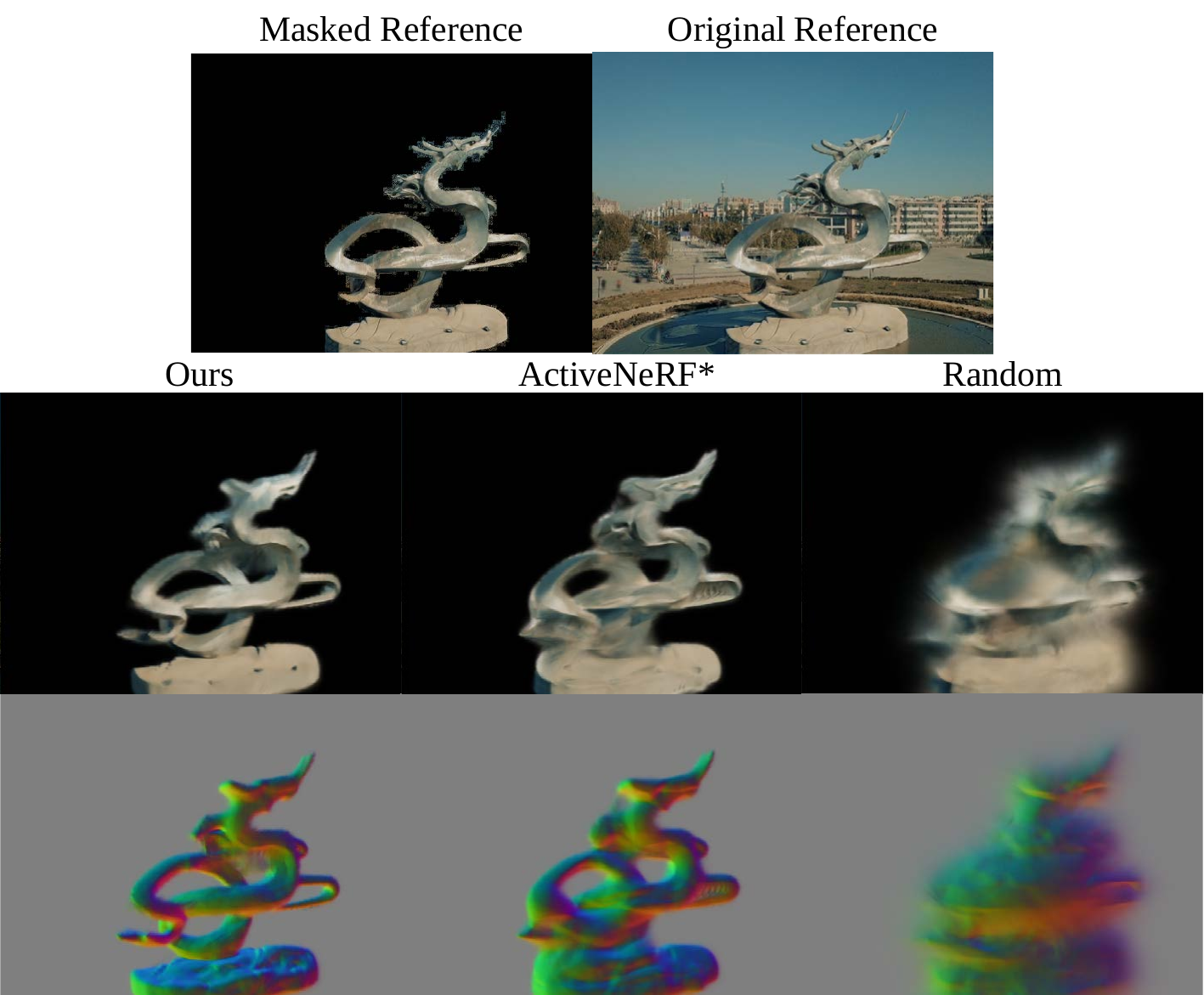}
     \caption{
     Rendered novel views and corresponding normal maps of three best-performing NBV selection methods using BlendedMVS dataset in ten-image setting. 
     First, second, and third rows show reference images of dragon, rendered novel views, and corresponding normal maps, respectively.
}
\label{fig:blendedMVS}
 \end{figure}

For the BlendedMVS dataset, we qualitatively compared our method with the second- and third-best performing NBV methods, as illustrated in \cref{fig:blendedMVS}. 
The dragon scene presented a significant challenge for the random method due to its complex shape. The random method failed to select views that contributed to reconstructing accurate normals, resulting in degraded RGB rendering outcomes.

ActiveNeRF* performed better than the random method by leveraging the color variations of the dragon caused by sunlight reflections and shadows. 
 By utilizing color uncertainty for view selection, ActiveNeRF* was able to capture these color variations more effectively. 
 However, it still failed to fully reconstruct the overall shape of the dragon, especially in the lower sections, which featured relatively simple colors.

In contrast, our method prioritized areas where the surface was yet to be reconstructed rather than focusing solely on already existing surfaces. This approach aligns with the principle that surface refinement should follow the completion of shape reconstruction.
As a result, our method successfully reconstructed the overall shape of the dragon more effectively, accurately identifying and filling gaps, such as holes in the dragon statue.

\begin{table}[!t]
  \centering
  \caption{Evaluation of mesh reconstruction on DTU
  }
  \begin{tabular}{l c c c}
Methods&Acc.$\downarrow$&Comp.$\downarrow$&Chamfer$\downarrow$\\
 \midrule

Random&\cellcolor{tabthird}2.368&\cellcolor{tabthird}3.472&\cellcolor{tabthird}2.920\\
FVS&3.425&4.411&3.918\\
Entropy&2.861&4.426&3.644\\
ActiveNeRF*&\cellcolor{tabsecond}2.026&\cellcolor{tabsecond}2.763&\cellcolor{tabsecond}2.395\\
FisherRF*&3.073&3.878&3.476\\
Ours&\cellcolor{tabfirst}1.829&\cellcolor{tabfirst}2.176&\cellcolor{tabfirst}2.002\\
  \end{tabular}
   \label{tab:mesh_dtu}
\end{table}

Based on the mesh reconstruction evaluation results presented in~\cref{tab:mesh_dtu}, our method outperformed all other methods on the DTU dataset. 
This indicates that our NBV selection method effectively prioritizes views that significantly contribute to mesh reconstruction quality.

The entropy method, which evaluates the entropy of occupancy probabilities, performed poorly in mesh reconstruction, contrary to expectations.
Referring to the IG formulation (see~\cref{dens_ent}), the entropy of an unknown voxel is higher than that of an occupied or vacant voxel. 
As training progresses and no unknown voxels remain, the focus shifts to voxels with median occupancy probability value among all voxels.
However, the entropy method does not account for surface occlusion. 
Consequently, it failed to distinguish between visible ambiguous regions and those that occluded behind surfaces. This limitation led to poor performance in both mesh reconstruction and image rendering.

\textbf{One- and four-image settings.} 
We present quantitative results in three types of experimental settings of NBV selection for the Blender dataset in \cref{tab:select_blender}.
Our method outperformed other NBV selection techniques in the two-image setting and demonstrated comparable performance to the best-performing method in the one-image setting.
When views are sparse, it is crucial to cover the entire scene while ensuring the convergence of the training model.
FVS, despite being a simple method that selects candidate views furthest from the training views, performed well in the one-image setting using the Blender dataset.
This effectiveness can be attributed to the characteristic of camera views in the Blender dataset being distributed in a hemisphere, with the target captured in the center of images without occlusion.

In the four-image setting, the FVS, FisherRF, and random methods with density representation outperformed our approach in image rendering criteria.
However, as shown in \cref{fig:blender_20_supple}, the mesh reconstruction of the best-performing method, FVS, is poor, as the model with density representation does not infer an explicit surface.
Interestingly, when the number of view selections in experimental settings increased from 2 to 4, the performance of methods using density representations improved significantly (\cref{tab:select_blender}).
This observation suggests a characteristic of density representation: it converges better in detail when the number of images exceeds a certain threshold.

 \begin{table*}[!t]
  \centering
  \caption{Evaluation of image rendering and mesh reconstruction on DTU. 
  }
  \begin{adjustbox}{width=1.0\linewidth,center}
  \begin{tabular}{l c c c c c c c c c c c c}
   &\multicolumn{6}{c}{two-image setting}&\multicolumn{6}{c}{four-image setting}\\
   \cmidrule(lr){2-7}
   \cmidrule(lr){8-13}
Methods&PSNR$\uparrow$&SSIM$\uparrow$&LPIPS$\downarrow$&Acc.$\downarrow$&Comp.$\downarrow$&Chamfer$\downarrow$&PSNR$\uparrow$&SSIM$\uparrow$&LPIPS$\downarrow$&Acc.$\downarrow$&Comp.$\downarrow$&Chamfer$\downarrow$\\
 \midrule
 \textbf{Density (except Ours)} \\
 \midrule
Random&\cellcolor{tabthird}21.17&\cellcolor{tabthird}0.769&\cellcolor{tabthird}0.238&\cellcolor{tabthird}5.166&8.680&\cellcolor{tabthird}6.923
&28.14&0.896&\cellcolor{tabthird}0.101&\cellcolor{tabsecond}5.084&\cellcolor{tabthird}5.757&\cellcolor{tabthird}5.420\\
FVS&18.95&0.730&0.268&6.149&9.822&7.986
&27.60&0.889&0.111&7.007&5.916&6.461\\
Entropy&16.86&0.684&0.299&\cellcolor{tabsecond}4.760&\cellcolor{tabsecond}6.886&\cellcolor{tabsecond}5.823&26.09&0.866&0.127&\cellcolor{tabthird}5.230&\cellcolor{tabsecond}5.580&\cellcolor{tabsecond}5.405\\
ActiveNeRF&\cellcolor{tabsecond}21.78&\cellcolor{tabsecond}0.786&\cellcolor{tabsecond}0.229&5.862&9.955&7.909&\cellcolor{tabsecond}29.56&\cellcolor{tabfirst}0.911&\cellcolor{tabfirst}0.090&6.345&8.269&7.307\\
FisherRF&16.50&0.737&0.318&6.884&\cellcolor{tabthird}7.925&7.405&\cellcolor{tabthird}28.97&\cellcolor{tabthird}0.903&\cellcolor{tabsecond}0.097&6.901&5.885&6.393\\
Ours&\cellcolor{tabfirst}28.19&\cellcolor{tabfirst}0.867&\cellcolor{tabfirst}0.168&\cellcolor{tabfirst}1.829&\cellcolor{tabfirst}2.176&\cellcolor{tabfirst}2.002&\cellcolor{tabfirst}30.10&\cellcolor{tabsecond}0.910&0.116&\cellcolor{tabfirst}1.715&\cellcolor{tabfirst}1.967&\cellcolor{tabfirst}1.864\\
\midrule
\textbf{Surface} \\
\midrule
Random&\cellcolor{tabthird}27.69&\cellcolor{tabsecond}0.864&\cellcolor{tabsecond}0.170&\cellcolor{tabthird}2.368&\cellcolor{tabthird}3.472&\cellcolor{tabthird}2.920
&\cellcolor{tabthird}29.34&\cellcolor{tabthird}0.907&\cellcolor{tabsecond}0.111&\cellcolor{tabsecond}1.765&\cellcolor{tabthird}2.340&\cellcolor{tabthird}2.053\\
FVS&27.09&0.852&0.187&3.425&4.411&3.918
&29.03&0.880&0.153&2.559&3.250&2.904\\
Entropy&24.21&0.810&0.218&2.861&4.426&3.644&27.36&0.872&0.147&\cellcolor{tabthird}2.206&3.198&2.702\\
ActiveNeRF*&26.30&0.852&\cellcolor{tabthird}0.175
&\cellcolor{tabsecond}2.026&\cellcolor{tabsecond}2.763&\cellcolor{tabsecond}2.395&28.25&0.873&0.137&2.702&\cellcolor{tabsecond}2.182&\cellcolor{tabsecond}2.021\\
FisherRF*&\cellcolor{tabsecond}27.78&\cellcolor{tabthird}0.860&0.180&3.073&3.879&3.476&\cellcolor{tabfirst}30.80&\cellcolor{tabfirst}0.916&\cellcolor{tabfirst}0.108&2.465&2.933&2.699\\
Ours&\cellcolor{tabfirst}28.19&\cellcolor{tabfirst}0.867&\cellcolor{tabfirst}0.168&\cellcolor{tabfirst}1.829&\cellcolor{tabfirst}2.176&\cellcolor{tabfirst}2.002&\cellcolor{tabsecond}30.10&\cellcolor{tabsecond}0.910&\cellcolor{tabthird}0.116&\cellcolor{tabfirst}1.715&\cellcolor{tabfirst}1.967&\cellcolor{tabfirst}1.864\\
  \bottomrule
  \end{tabular}
  \end{adjustbox}
   \begin{minipage}{\textwidth}
  \vspace{0.1cm}
      \small
The mesh reconstruction of the reference point clouds was performed using the method proposed by Kazhdan et al. \cite{kadzhdan2006poisson}.
For evaluation, the Chamfer distance metric in MonoSDF was used \cite{Yu2022MonoSDF}.
  \end{minipage}
   \label{tab:select_dtu}
\end{table*}

\begin{figure*}[!t]
\centering
\subfloat[
\small
Comparisons of the two best-performing with the surface representations on DTU scan 106.]{\includegraphics[width=1.\columnwidth]{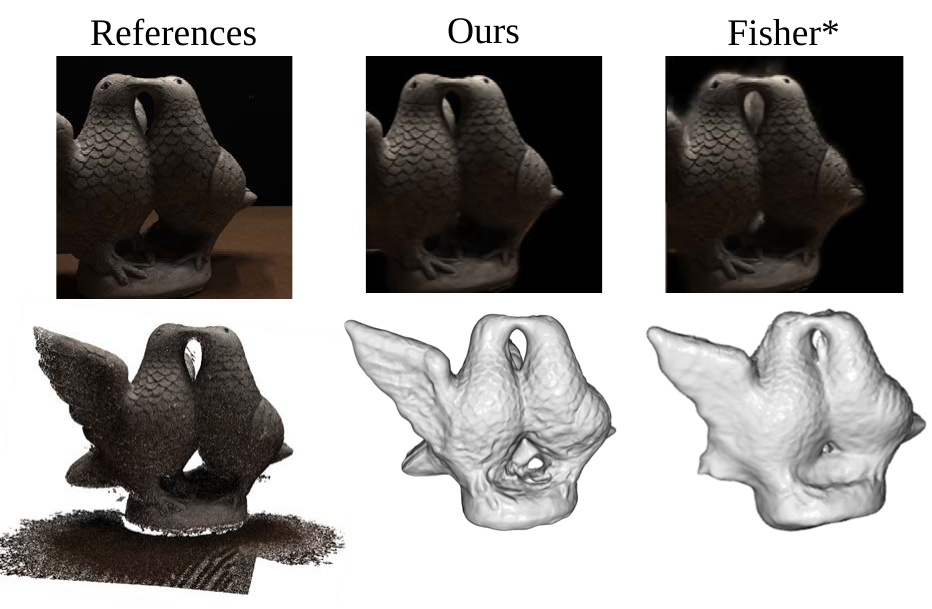}%
\label{fig:dtu_20_106supple}}
\hfil
\subfloat[
\small
Comparisons of the best-performing with the density representation and ours on DTU scan 97.]{\includegraphics[width=0.9\columnwidth]{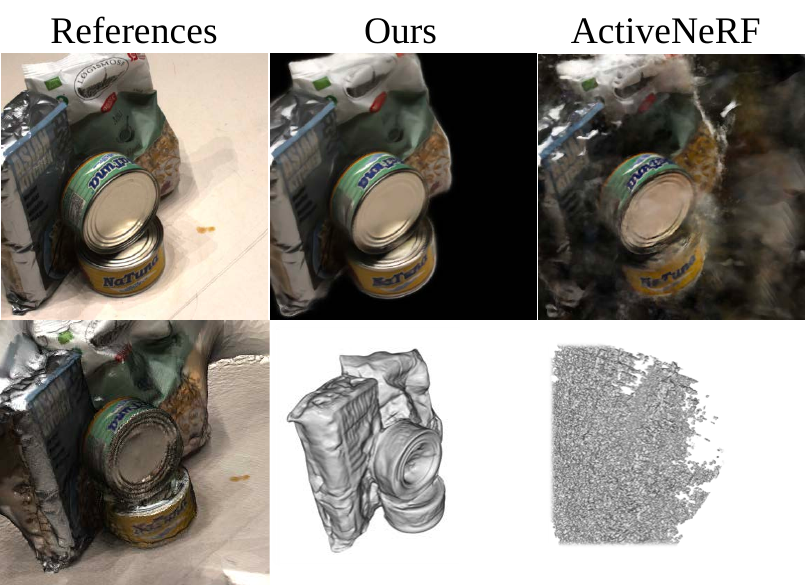}%
\label{fig:dtu_20_supple}}
\hfill
\caption{Rendered images and reconstructed meshes of the compared methods in a four-image setting. The second row of the references shows the ground truth point cloud instead of a mesh.
}
\label{fig:dtu_supple}
\end{figure*}

In the DTU dataset (\cref{tab:select_dtu}), 
the FisherRF* in the four-image setting outperforms our method in image rendering criteria. Nevertheless, as shown in \cref{fig:dtu_20_106supple}, the image rendering near the object surface is slightly blurred, and the mesh reconstruction lacks detailed features.
The ActiveNeRF with density representation in the four-image setting performed comparably to our approach in image rendering criteria.
However, as illustrated in \cref{fig:dtu_20_supple}, ActiveNeRF exhibited poor mesh reconstruction results.
Moreover, ActiveNeRF rendered cloudy artifacts outside the object masks, where the training gradient force from color loss is masked out.

Interestingly, in density representation, ActiveNeRF demonstrates the second-best image rendering performance in two- and four-image settings, while the entropy method shows the second-best mesh reconstruction performance in the same settings.
These results indicate that each method excels in the criteria aligned with their IG formulation:
ActiveNeRF focuses on quantifying IG based on color uncertainty, whereas the entropy method emphasizes IG based on occupancy probabilities.
However, this trend is not observed in the surface representation.
Instead, ActiveNeRF* enhances mesh reconstruction performance by incorporating the Eikonal loss during model training, even when views are selected based on color uncertainty.
In contrast, the entropy method in surface representation does not excel in mesh reconstruction, as the Eikonal loss from NeuS has generally improved this aspect of performance across all methods. To outperform other methods in mesh reconstruction criteria, the entropy method should consider the visibility of uncertain regions in its view selection process.

In the DTU dataset, the FVS method does not demonstrate superior performance, unlike in the Blender dataset.
The DTU dataset comprises images of target objects captured by a robot arm moving within a limited region, resulting in camera views distributed within the arm's reach.
Furthermore, the camera views only cover the front side of the target objects, which may cause the FVS method to select views positioned more at the boundary. This bias can potentially lead the FVS method to miss crucial front-side details.

\begin{figure}[!t]
     \centering
\includegraphics[width=1.0\linewidth]{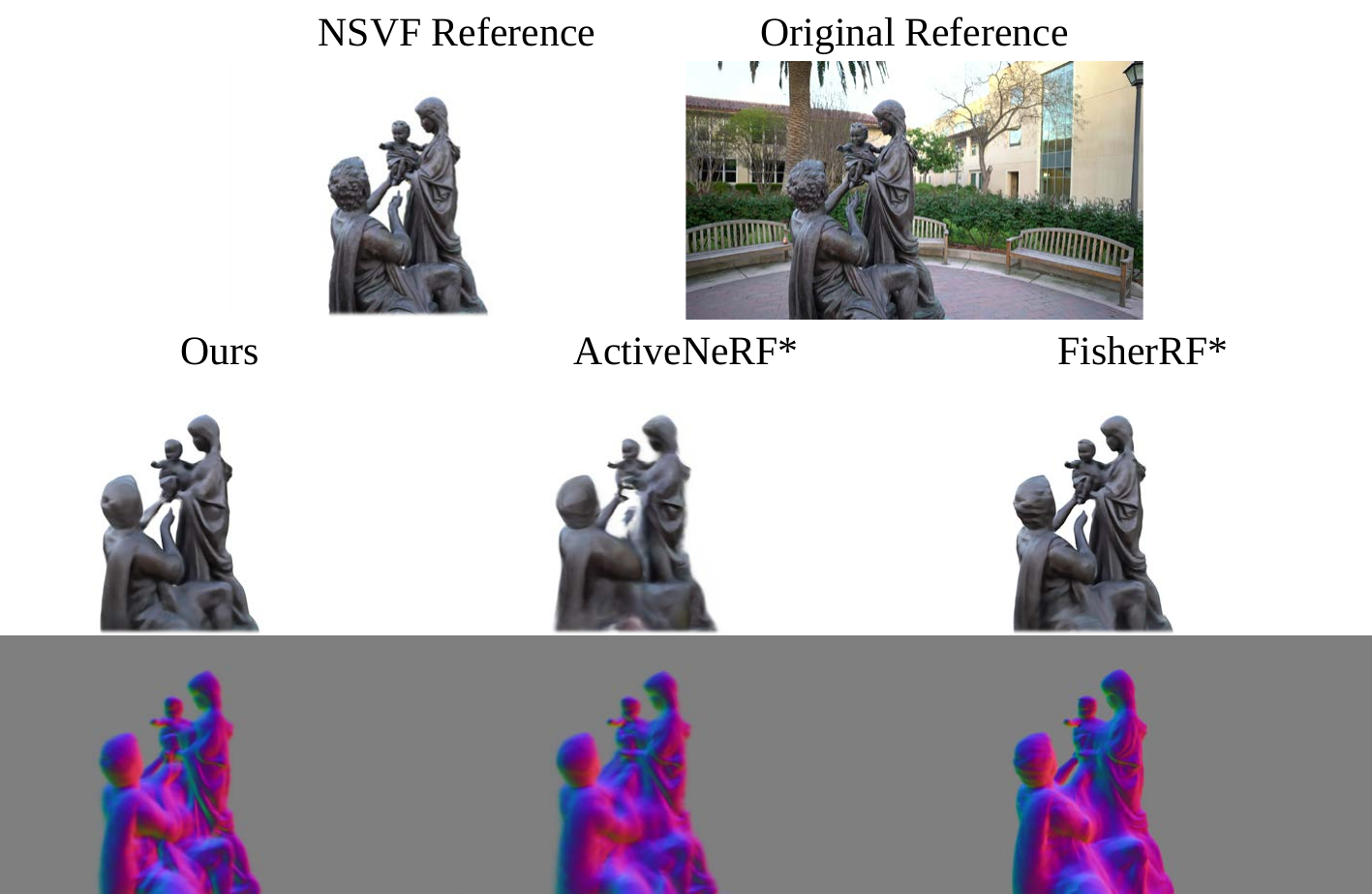}
     \caption{
 Rendered images and estimated normals of the three best-performing NBV selection methods in the TanksAndTemples (Family) dataset.
}
\label{fig:tnt_family}
 \end{figure}

In the TanksAndTemples dataset (\cref{tab:select_tnt}), our method outperforms other approaches in 3 out of 5 scenes.
For the family scene, while our approach yields a lower PSNR compared to FisherRF*, it still effectively renders and reconstructs the scene, as illustrated in \cref{fig:tnt_family}. FisherRF*, however, captures finer details in clothing folds and hair curls.

\begin{figure}[!t]
     \centering
\includegraphics[width=1.0\linewidth]{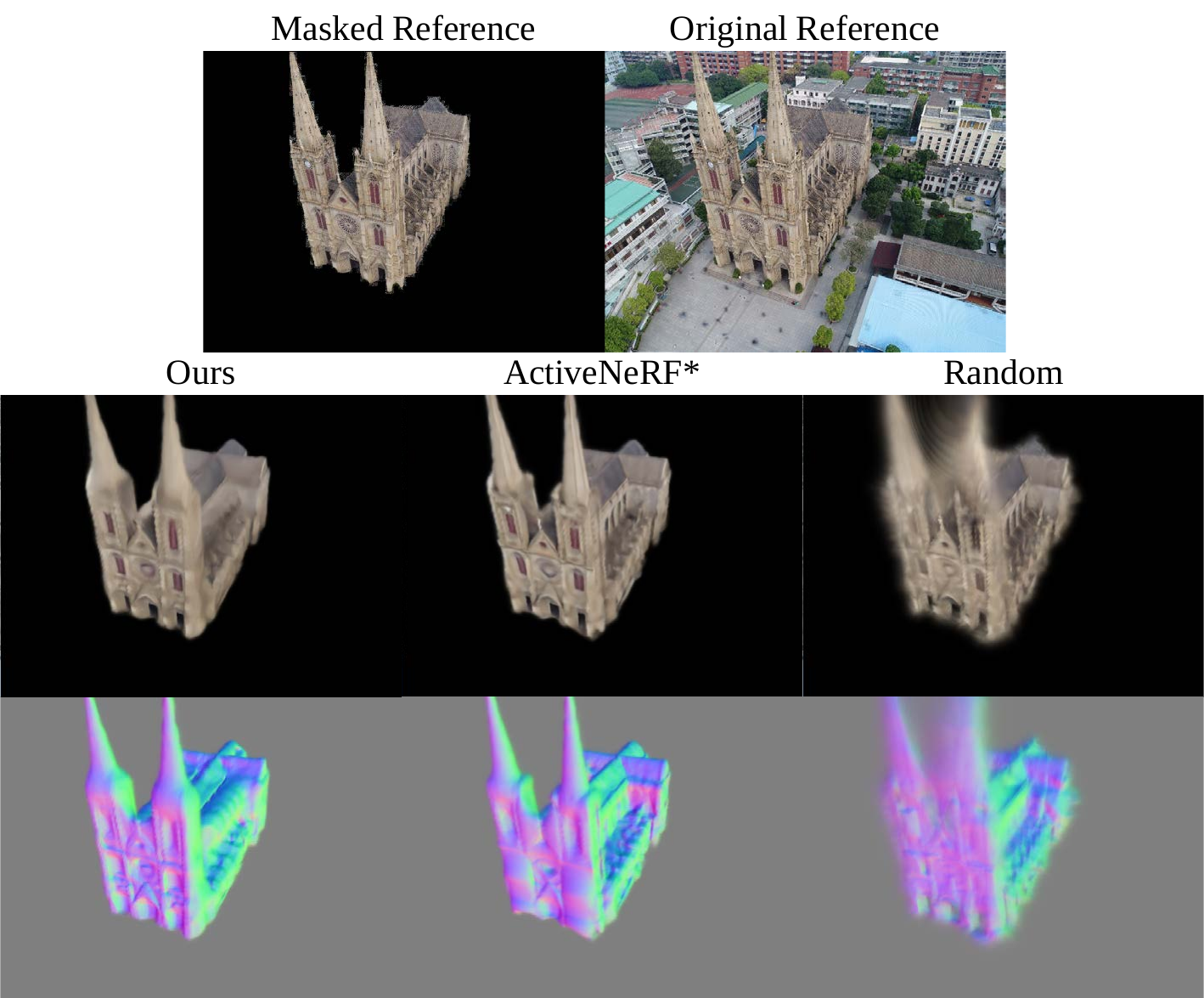}
     \caption{
 Rendered images and estimated normals of the three best-performing NBV selection methods in the BlendedMVS (Cathedral) dataset.
}
\label{fig:blended_cathedral}
 \end{figure}

In the BlendedMVS dataset (\cref{tab:select_mvs}), our method outperforms others on average and achieves the best results in two scenes.
For the cathedral scene, our method demonstrates the third-highest PSNR score, while ActiveNeRF* and the random method show the best and second-best performances, respectively.
However, the 3D reconstruction results of each method exhibit both strengths and weaknesses, reflecting the characteristics of each approach.
As shown in \cref{fig:blended_cathedral}, ActiveNeRF* renders the cathedral's front view with detailed red blocks, but its normal estimation for this view lacks fine details.
The random method successfully renders and reconstructs the cathedral's side section, but it demonstrates poor performance in rendering and reconstructing the cathedral's steeples.
Furthermore, the image rendered using the random method fails to capture the red color in the front view.
Our method effectively renders and reconstructs the front view, but it produces blurred images of the cathedral's side section and exhibits low convergence in mesh reconstruction, resulting in a lower estimated PSNR.

\begin{table}[!t]
  \centering
  \caption{
 PSNR comparison on the TanksAndTemples. 
   }
  \label{tab:select_tnt}
  \begin{adjustbox}{width=1.0\linewidth,center}
  \begin{tabular}{l c c c c c c}
Scene&Random&FVS&Entropy&ActiveNeRF*&FisherRF*&Ours\\
\midrule
Barn       &12.76&\cellcolor{tabthird}13.2&12.73&12.17&\cellcolor{tabsecond}13.87&\cellcolor{tabfirst}17.98\\
Caterpillar&11.14&11.95&10.39&\cellcolor{tabsecond}17.54&\cellcolor{tabthird}12.72&\cellcolor{tabfirst}17.77\\
Ignatius   &\cellcolor{tabsecond}19.81&\cellcolor{tabthird}19.57&18.05&18.59&\cellcolor{tabfirst}22.17&17.77\\
Truck      &12.86&\cellcolor{tabthird}15.2&13.95&\cellcolor{tabsecond}18.77&13.38&\cellcolor{tabfirst}19.95\\
Family     &\cellcolor{tabthird}29.26&\cellcolor{tabsecond}29.8&28.4&26.01&\cellcolor{tabfirst}30.07&29\\
\bottomrule
Mean       &17.17&17.94&16.70&\cellcolor{tabsecond}18.62&\cellcolor{tabthird}18.44&\cellcolor{tabfirst}20.49\\
  \end{tabular}
  \end{adjustbox}
\end{table}

 \begin{figure}
     \centering
      \includegraphics[width=1.\linewidth]{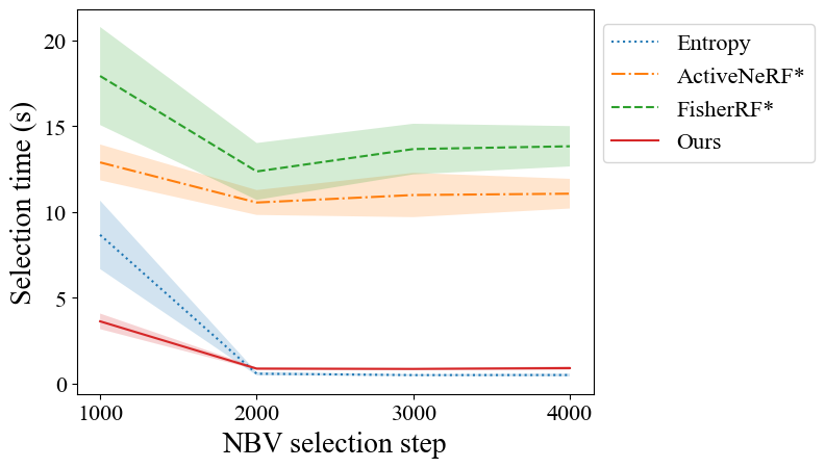}
     \caption{
     Time comparison for NBV selection using the Blender dataset in a two-image setting. The line graph shows the average over eight scenes, with the shaded area representing the standard deviation. 
}
\label{fig:nbv_select_time}
 \end{figure}
 
\textbf{Time comparison.} We evaluated the time required to select the NBVs in the two-image setting using the Blender dataset(\cref{fig:nbv_select_time}). Our method (0.8 s) significantly reduced time consumption compared to ActiveNeRF* (10.8 s) and FisherRF* (13.5 s). It also exhibited similar time efficiency to the entropy method (0.5 s), which is likewise based on voxel-based IG computation. 
FisherRF* displayed increased time consumption compared to its original implementation \cite{Jiang2023FisherRF}. 
This is primarily due to a shift in the base model architecture from 3D Gaussians \cite{kerbl3Dgaussians} to NeuS \cite{wang2021neus}.
While 3D Gaussians, as explicit representations, reduce computation time, they are constrained by a quality-memory trade-off, posing implementation challenges on the RTX2080 GPU.
Additionally, this change in model architecture was necessary to ensure a fair comparison on the IG computation methods within the same architecture framework, highlighting the efficiency of our approach.

\section{New Dataset with Imbalanced Viewpoints}
 \label{new_dataset}

\begin{figure*}[!t]
     \centering
\includegraphics[width=0.9\linewidth]{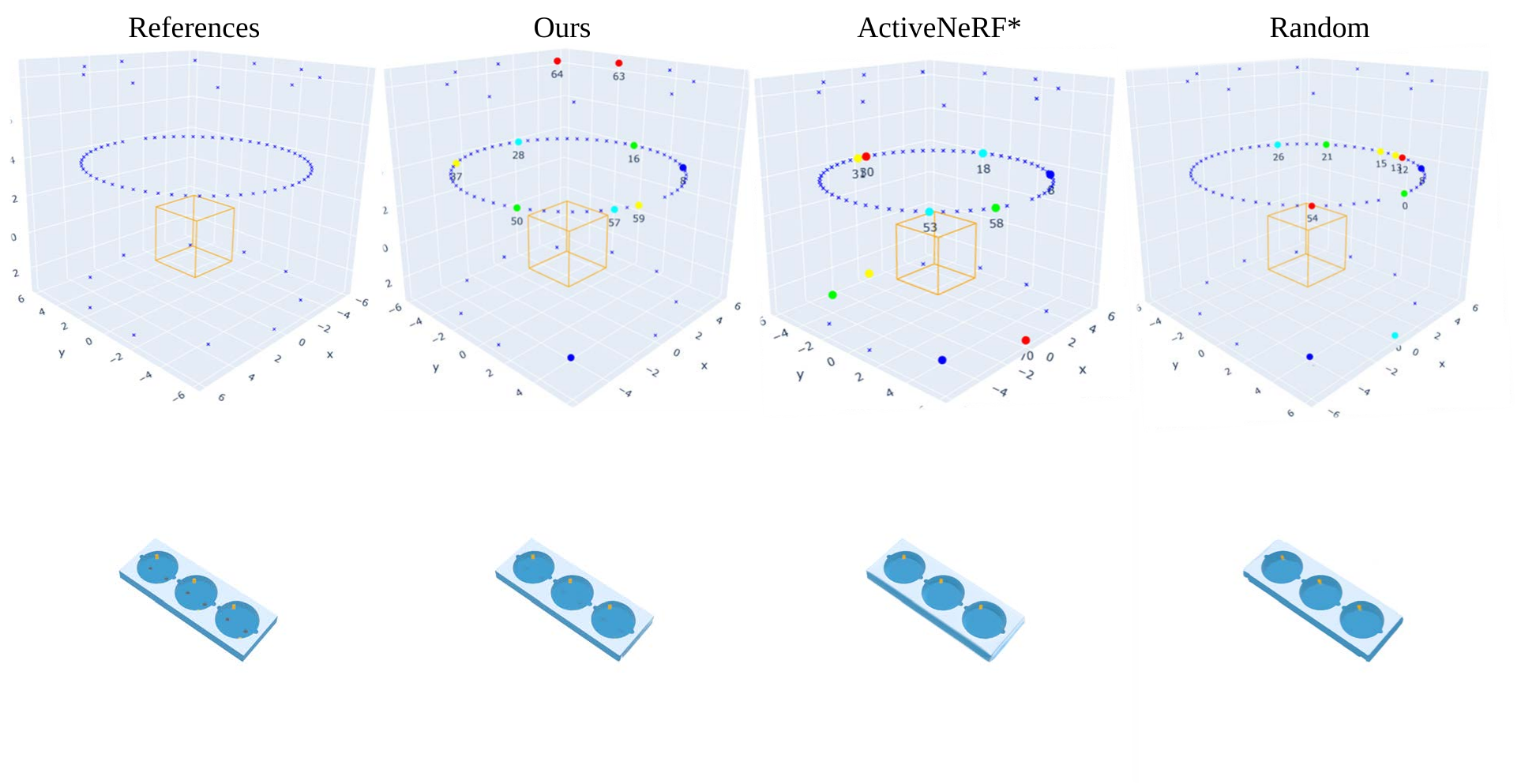}
     \caption{
     View selections and rendered images of the three best-performing NBV selections in two-image settings using the ImBView dataset are compared. 
     In the two-image setting, two NBVs are selected at each view-selection iteration. The results of the view selection are shown in the order of blue-cyan-green-yellow-red.
     The initial two views (shown in blue) are fixed in all NBV selection methods.
}
\label{fig:custom_outlet_view}
 \end{figure*}

We present the view selections and novel view synthesis results for the outlet scene in the ImBView dataset. The quantitative results are reported in \cref{tab:psnr_custom} of the main manuscript.
As illustrated in \cref{fig:custom_outlet_view}, the random and ActiveNeRF* methods predominantly selected views from common viewpoints positioned in the middle height, resulting in their failure to reconstruct the two black holes in each outlet.
In contrast, our method successfully reconstructs these two black holes (albeit with some color inaccuracies) through strategic view selection, distributed across three categories of viewpoints: high-angle, common, and low-angle perspectives.

  \begin{figure*}
     \centering
      \includegraphics[width=0.9\linewidth]{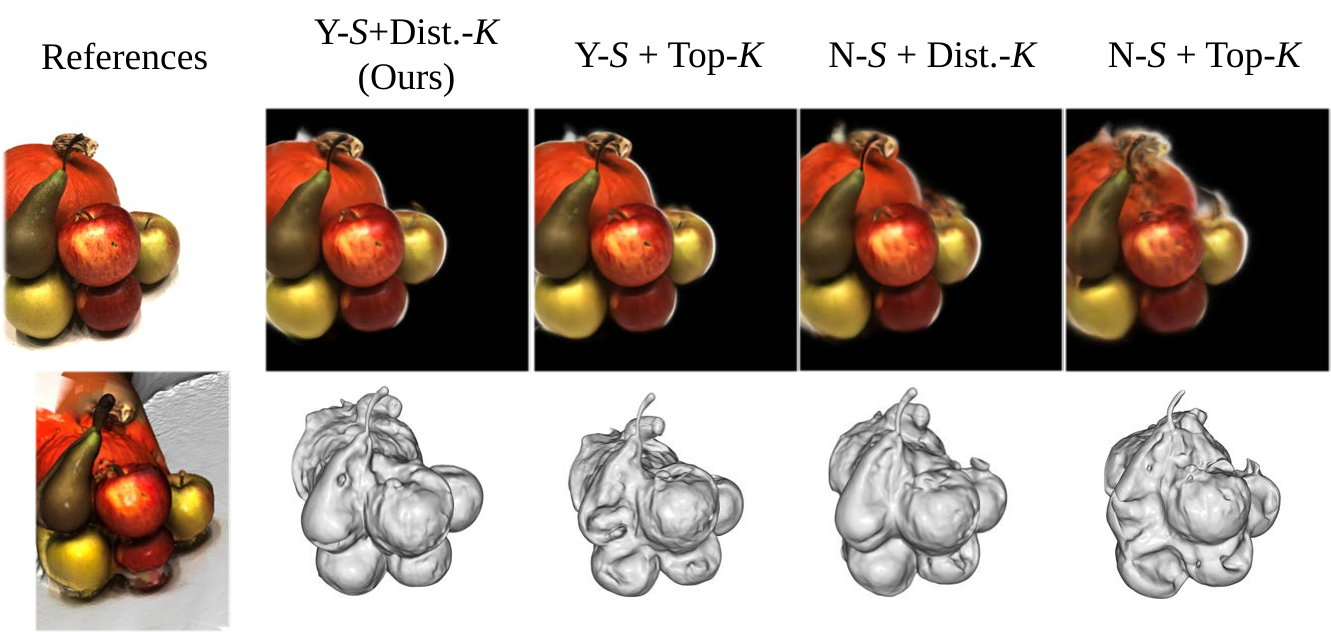}
     \caption{
     Rendered RGB images and reconstructed meshes are shown in the first and second rows.
     The ablation studies using the DTU dataset are conducted in a two-image setting.
}
\label{fig:dtu_ablation}
 \end{figure*}

  \begin{figure*}
     \centering
      \includegraphics[width=0.9\linewidth]{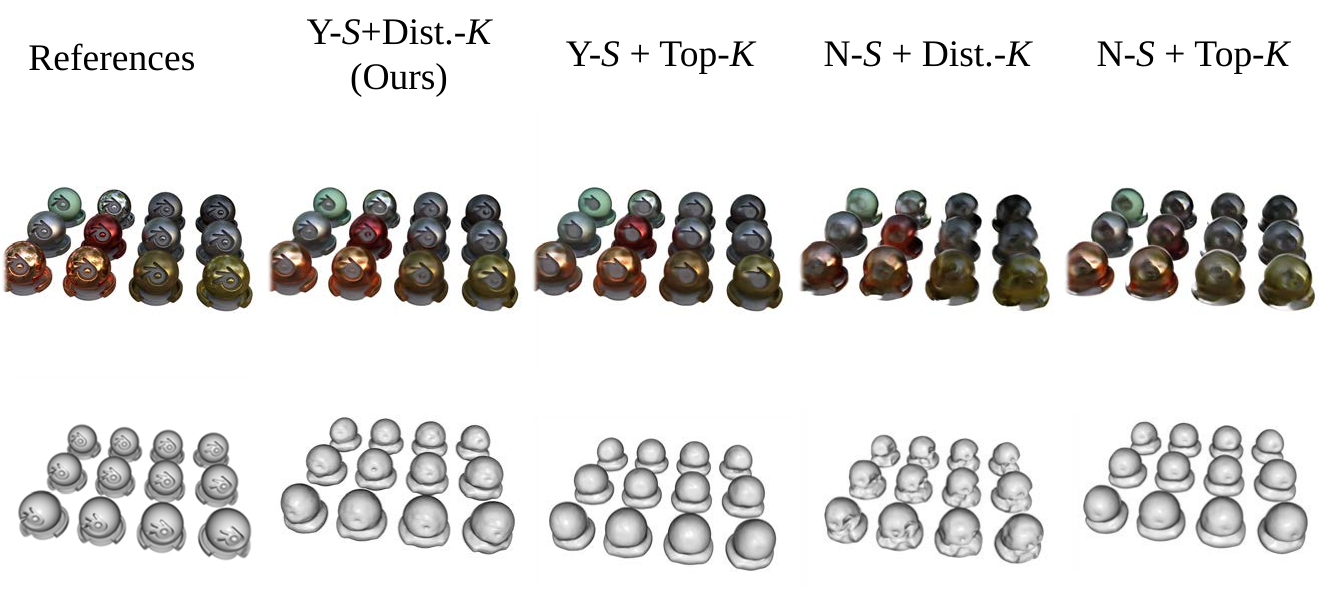}
     \caption{
     Rendered RGB images and reconstructed meshes are shown in the first and second rows.
     The ablation studies using the Blender dataset are conducted in a two-image setting.
}
\label{fig:blender_ablation}
 \end{figure*}

\section{Qualitative Results on Ablation Studies}
\label{append:multiple}

\begin{table*}[!t]
\caption{Ablation studies on SBV-based view selection in two-image setting.  
  }
  \centering
  \begin{adjustbox}{width=1.0\linewidth,center}
    \begin{tabular}{c c c c c c c c c c c}
   &&\multicolumn{3}{c}{Blender}&\multicolumn{6}{c}{DTU}\\
   \cmidrule(lr){3-5}
 \cmidrule(lr){6-11}
  \textit{S}&$K$ &PSNR$\uparrow$&SSIM$\uparrow$&LPIPS$\downarrow$&PSNR$\uparrow$&SSIM$\uparrow$&LPIPS$\downarrow$&Acc.$\downarrow$&Comp.$\downarrow$&Chamfer$\downarrow$\\
  \cmidrule(lr){1-2}
 \cmidrule(lr){3-5}
 \cmidrule(lr){6-8}
 \cmidrule(lr){9-11}
\multirow{2}{*}{Y}&Dist.&\textbf{21.22}&\textbf{0.854}&\textbf{0.170}&\textbf{28.19}&\textbf{0.867}&\textbf{0.168}&\textbf{1.829}&\textbf{2.176}&\textbf{2.002}\\
&Top&20.03&0.843&0.185&27.56&0.858&0.174&2.041&2.508&2.274\\
\midrule
\multirow{2}{*}{N}&Dist.&20.13&0.844&0.181&27.18&0.855&0.180&2.060&2.441&2.251\\
&Top&19.94&0.842&0.187&25.64&0.836&0.193&2.127&2.613&2.370\\
  \end{tabular}
  \end{adjustbox}
  \label{tab:ablation_topk}
\end{table*}

We present qualitative results for the ablation studies reported in \cref{tab:ablation_topk}.
In the DTU dataset (\cref{fig:dtu_ablation}), next-best view selection methods that incorporate surface confidence (Y-\textit{S}) effectively reconstructed the stem of the large apple at the top of the image and the yellow apple on the right.
The mesh reconstructions utilizing Dist.-$K$ methods accurately captured the convexity of each fruit.
Consequently, our SBV-based view selection (Y-\textit{S} + Dist.-$K$) outperformed other ablations, demonstrating the efficacy of surface-aware IG computation combined with distance-aware NBV selections.

\begin{table}[!t]
  \centering
  \caption{
 PSNR comparison on the BlendedMVS. 
   }
  \label{tab:select_mvs}
    \begin{adjustbox}{width=1.0\linewidth,center}
  \begin{tabular}{l c c c c c c}
Scene & Random & FVS & Entropy & ActiveNeRF* & FisherRF* & Ours \\
\hline
Cathedral & \cellcolor{tabsecond}23.29 & 20.96 & 21.82 & \cellcolor{tabfirst}23.88 & 20.94 & \cellcolor{tabthird}23.27 \\
Dragon & \cellcolor{tabthird}24.95 & 24.38 & 24.94 & \cellcolor{tabsecond}27.79 & 23.55 & \cellcolor{tabfirst}29.52 \\
Man & \cellcolor{tabsecond}34.97 & \cellcolor{tabthird}34.57 & 33.99 & 31.39 & \cellcolor{tabfirst}36.17 & 32.28 \\
Pavilion & 19.75 & \cellcolor{tabthird}20.77 & 19.78 & \cellcolor{tabfirst}21.72 & 19.94 & \cellcolor{tabsecond}20.79 \\
Red dome & \cellcolor{tabsecond}28.11 & 26.79 & 28.06 & \cellcolor{tabthird}28.07 & 27.97 & \cellcolor{tabfirst}28.12 \\
\hline
Mean & \cellcolor{tabthird}26.21 & 25.49 & 25.72 & \cellcolor{tabsecond}26.57 & 25.71 & \cellcolor{tabfirst}26.80 \\
\end{tabular}
  \end{adjustbox}
\end{table}

In the Blender dataset (\cref{fig:blender_ablation}), the Y-\textit{S} methods successfully rendered the light reflections in twelve ball-shaped objects.
The N-\textit{S} methods exhibited deteriorated color rendering, with the N-\textit{S}+Dist.-$K$ method particularly failing to render light reflections and inadequately reconstructing the lower right portion of the target objects.
The deteriorated color renderings in N-\textit{S} methods may be attributed to their focus on non-surface regions with high color uncertainties during early training stages.
As training progressed, these non-surface regions might have been determined to be empty spaces, resulting in missed opportunities to refine the reconstruction of object surfaces.
Notably, the distance-aware NBV selection strategies may have exacerbated this issue by reinforcing the selection of views that contribute less to image rendering.

\end{document}